\documentclass{article}

\PassOptionsToPackage{numbers, compress}{natbib}

\usepackage[preprint]{neurips_2026}

\usepackage[utf8]{inputenc} 
\usepackage[T1]{fontenc}    
\usepackage{hyperref}       
\usepackage{url}            
\usepackage{booktabs}       
\usepackage{amsfonts}       
\usepackage{nicefrac}       
\usepackage{microtype}      

\usepackage{amsfonts}
\usepackage{amsmath,amssymb,mathtools}
\usepackage{graphicx}
\usepackage{float}
\usepackage[table]{xcolor}
\usepackage{makecell}
\usepackage{algorithm}
\usepackage{algorithmic}
\usepackage{wrapfig}
\usepackage{multirow}

\usepackage{placeins}
\title{Frequency-Synchronized Boundary Coupling for Training-Free Multi-Prompt Long Video Generation}

\author{%
  \textbf{Yipan Xu}$^{1}$ \quad
  \textbf{Shuyong Gao}$^{2}$\thanks{Corresponding authors.} \quad
  \textbf{Jiyuan Fu}$^{1}$ \quad
  \textbf{Keliang Yin}$^{1}$ \quad
  \textbf{Jiayu Chen}$^{1}$ \\[0.1em]
  \textbf{Qianyu Guo}$^{3}$ \quad
  \textbf{Yang Liu}$^{4}$ \quad
  \textbf{Wenqiang Zhang}$^{1}$\footnotemark[1] \\[0.5em]
  {\normalfont\normalsize
  $^{1}$Shanghai Key Lab of Intelligent Information Processing,}\\
  {\normalfont\normalsize
  College of Computer Science and Artificial Intelligence, Fudan University}\\
  {\normalfont\normalsize
  $^{2}$Naval Medical Center of PLA, Second Military Medical University}\\
  {\normalfont\normalsize
  $^{3}$Shanghai Institute of Virology, Shanghai Jiao Tong University}\\
  {\normalfont\normalsize
  $^{4}$College of Electronic and Information Engineering, Tongji University}\\[0.6em]
  {\normalfont\small
  \textbf{Project page:} \url{https://yipanxu-22.github.io/FreqSync-project}}
}

\begin{document}

\maketitle

\vspace{-1.5em}

\begin{abstract}
Text-to-video generation has made remarkable progress with diffusion models and transformer-based video architectures, yet generating coherent long videos from evolving textual descriptions remains challenging. Training long-video models from scratch requires substantial computation and large-scale multi-event data, while training-free multi-prompt generation must activate new prompt semantics without breaking temporal continuity. We identify a frequency-agnostic boundary coupling problem in existing transition strategies: weak coupling causes abrupt changes, flicker, and background jumps, whereas overly strong or coarse coupling may over-preserve source semantics and produce ghosting or duplicated subjects around prompt boundaries. To address this problem, we propose \textsc{FreqSync}, a training-free framework that formulates prompt transitions as frequency-selective boundary coupling. \textsc{FreqSync} combines Frequency-Synchronized Source-Conditioned Attention (FS-SCA) with Residual-Clipped Spectral Stitching (RCSS), enabling adjacent segments to share stable low- and mid-frequency structure and motion cues while keeping high-frequency prompt-specific details target-dominant under explicit residual constraints. Extensive experiments show that \textsc{FreqSync} achieves state-of-the-art transition smoothness and prompt alignment while maintaining strong perceptual video quality, demonstrating its effectiveness for coherent multi-prompt long video generation.
\end{abstract}

\vspace{-1.4em}
\section{Introduction}
\label{sec:introduction}

Text-to-video generation has advanced rapidly with diffusion models and transformer-based video architectures. Representative systems, including Imagen Video, Make-A-Video, ModelScopeT2V, Stable Video Diffusion, LaVie, AnimateDiff, Lumiere, VideoCrafter, CogVideoX, Open-Sora, HunyuanVideo, Wan, and Movie Gen, have demonstrated strong short-video generation quality from a single prompt~\cite{ho2022imagen,singer2022make,wang2023modelscope,blattmann2023stable,wang2025lavie,guo2023animatediff,bar2024lumiere,chen2024videocrafter2,yang2024cogvideox,zheng2024open,kong2024hunyuanvideo,wan2025wan,polyak2024movie}. Recent diffusion transformer (DiT) and multi-modal diffusion transformer (MM-DiT) models further improve scalability and text-video modeling capacity~\cite{peebles2023scalable,esser2024scaling,ma2024latte}. However, real-world videos often involve multiple events, scene changes, camera motions, and evolving object states, which are more naturally described by a sequence of prompts rather than a single static description~\cite{villegas2022phenaki,lin2023videodirectorgpt,cai2025ditctrl}.

\begin{figure}[t]
    \centering
    \includegraphics[width=\linewidth]{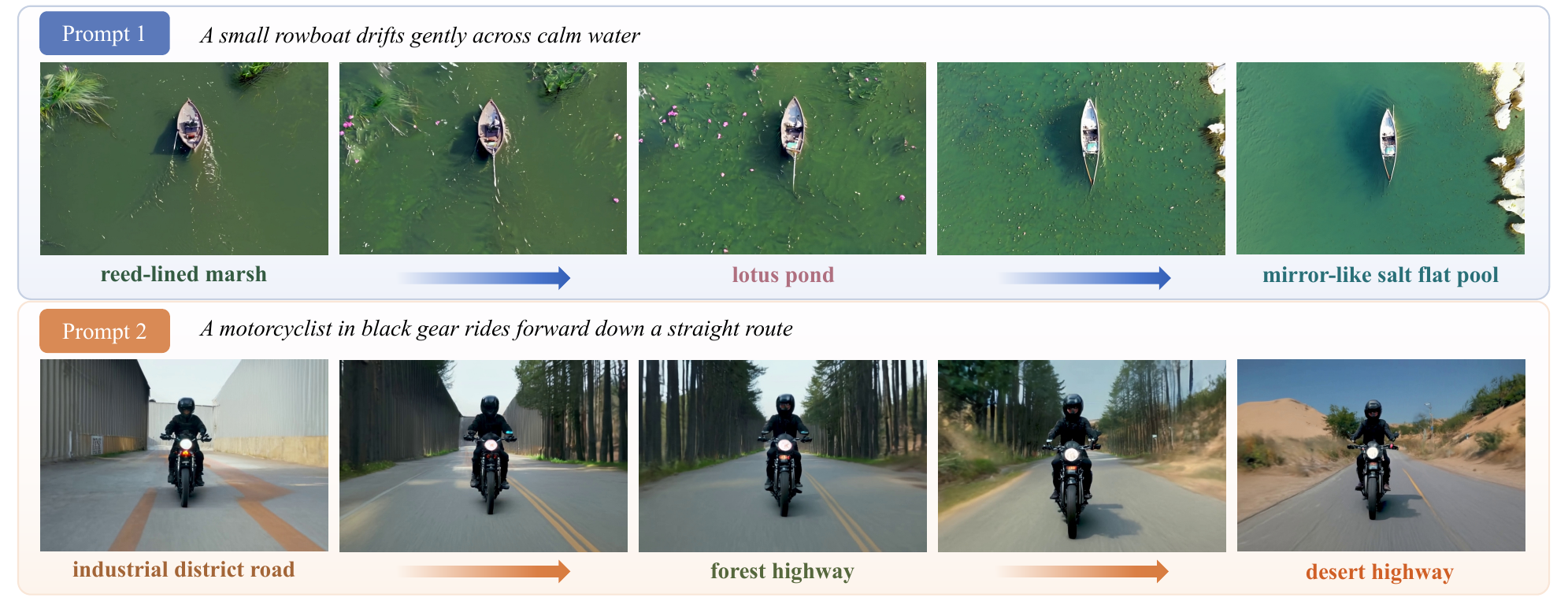}

    \caption{Teaser of \textsc{FreqSync}. Given a prompt sequence, \textsc{FreqSync} generates coherent long videos with smooth transitions, stable motion, and prompt-specific details.}
    \label{fig:teaser}
    \vspace{-1.0em}
\end{figure}

Multi-prompt long video generation requires each segment to follow its own prompt while maintaining smooth visual trajectories across prompt boundaries. Directly using a long prompt often causes weak prompt disentanglement, whereas independently generating and concatenating clips leads to abrupt transitions. Training long-horizon generators from scratch requires substantial computation and large-scale multi-event video-text data~\cite{brooks2024video,yang2024cogvideox,kong2024hunyuanvideo,wan2025wan,polyak2024movie}. This motivates training-free adaptation of pre-trained short-video generators to coherent multi-prompt long video generation.

\begin{figure}[t]
    \centering
    \includegraphics[width=0.95\linewidth]{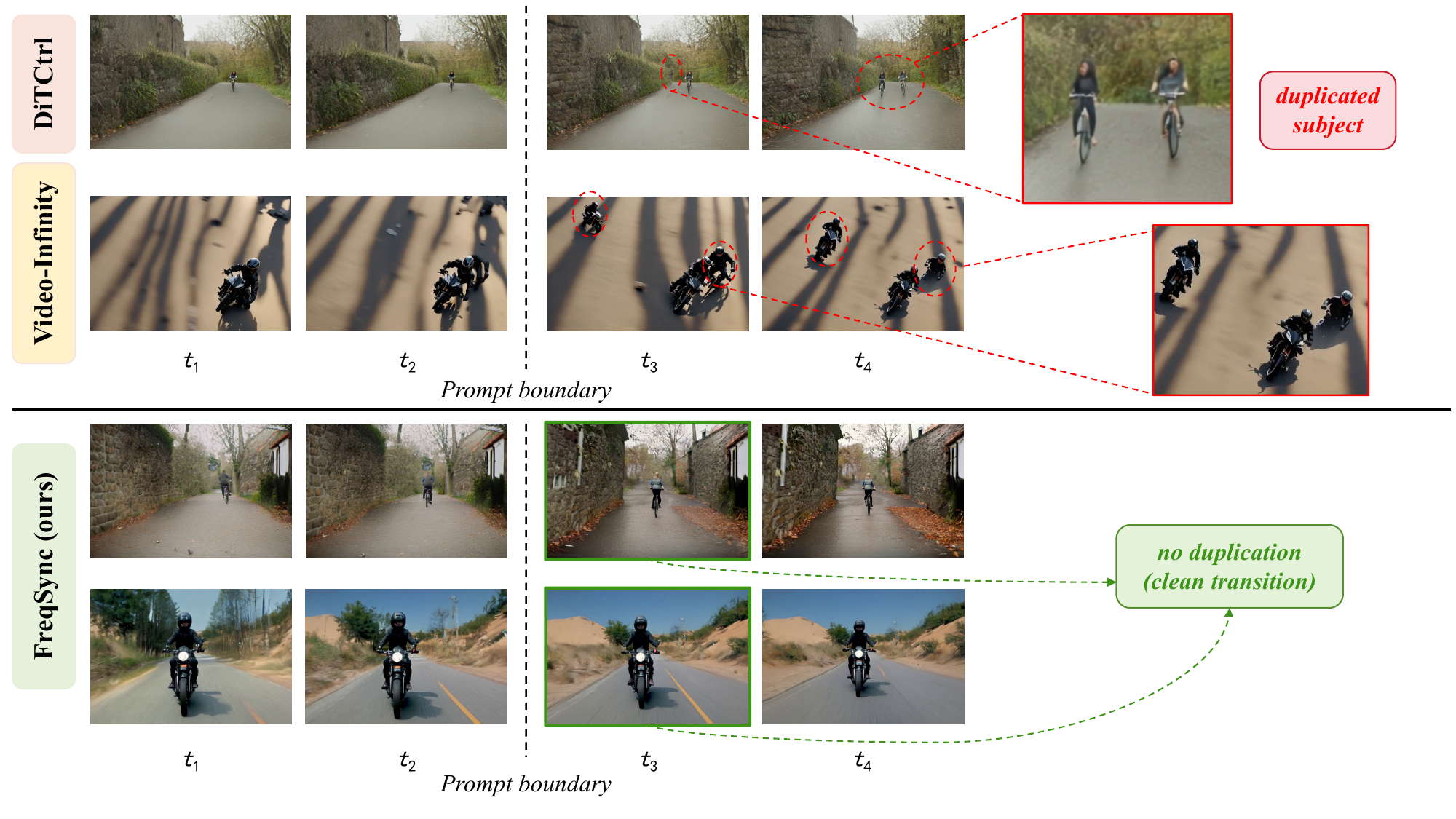}
    \vspace{-1.0em}
    \caption{Diagnostic observation. Strong frequency-agnostic coupling may over-preserve source-side evidence and cause duplicated subjects. \textsc{FreqSync} transfers mainly low- and mid-frequency continuity cues while keeping target details controlled.}
    \label{fig:diagnostic_overconditioning}

    \vspace{-1.0em}
\end{figure}

Existing training-free long-video methods improve temporal extension through noise rescheduling, sliding-window attention, temporal co-denoising, distributed inference, or frequency-domain blending~\cite{wang2023gen,qiu2023freenoise,kim2024fifo,tan2024video,lu2024freelong,li2025longdiff}. For multi-prompt generation, DiTCtrl introduces cross-segment KV-sharing and latent blending, showing that adjacent segments must be coupled for smooth prompt transitions~\cite{cai2025ditctrl}. However, our diagnostic observation suggests that coarse or frequency-agnostic coupling is insufficient. Weak coupling causes flicker, background jumps, and discontinuous motion, while overly strong coupling may over-preserve source-side layouts and semantics. As shown in Figure~\ref{fig:diagnostic_overconditioning}, such source retention can produce ghosting or duplicated subjects near boundaries.

We summarize this issue as a frequency-agnostic boundary coupling problem. A prompt boundary should not be handled by full-frequency source reuse or coarse overlap blending alone. Instead, source-side low- and mid-frequency components should provide layout, illumination, and motion continuity, while target-side high-frequency components should preserve newly introduced prompt details. This decomposition should guide both attention-level inheritance and prediction-level stitching.

To this end, we propose \textsc{FreqSync}, a training-free framework for multi-prompt long video generation. \textsc{FreqSync} formulates prompt transitions as frequency-selective boundary coupling and introduces two complementary modules. Frequency-Synchronized Source-Conditioned Attention (FS-SCA) performs frequency-selective source conditioning inside selected attention layers, enabling the target segment to inherit boundary-relevant structure without copying full-frequency source features. Residual-Clipped Spectral Stitching (RCSS) operates on denoised overlap predictions, performing spectral seam alignment and conservative low-frequency continuation under residual constraints. Together, FS-SCA and RCSS improve cross-segment continuity while reducing semantic contamination and stitching artifacts. Figure~\ref{fig:teaser} shows representative generated results.

Our contributions are summarized as follows:
\begin{itemize}

    \item We identify a frequency-agnostic boundary coupling problem in training-free multi-prompt long video generation. Weak coupling leads to discontinuous transitions, whereas overly strong or coarse coupling may over-preserve outdated source semantics, duplicate subjects, and introduce boundary artifacts.

    \item We propose \textsc{FreqSync}, a training-free frequency-synchronized framework for prompt-boundary coupling. It integrates Frequency-Synchronized Source-Conditioned Attention (FS-SCA) for frequency-selective attention-level source conditioning and Residual-Clipped Spectral Stitching (RCSS) for residual-constrained prediction-level boundary stitching.

    \item Extensive experiments on a 130-case multi-prompt benchmark demonstrate that \textsc{FreqSync} achieves the strongest overall performance, improving transition smoothness, temporal consistency, prompt alignment, and perceptual video quality over representative multi-prompt and long-video generation baselines.

\end{itemize}

\section{Related Work}
\label{sec:related_work}

\paragraph{Text-to-video diffusion models.}
Diffusion models have become the dominant paradigm for visual generation~\cite{ho2020denoising,song2020denoising,rombach2022high}. Early and representative video generators extend diffusion modeling to the temporal domain through latent video diffusion, motion modules, cascaded generation, or space-time architectures~\cite{ho2022video,singer2022make,ho2022imagen,blattmann2023align,zhou2022magicvideo,wang2023modelscope,guo2023animatediff,chen2023videocrafter1,blattmann2023stable,wang2025lavie,bar2024lumiere,chen2024videocrafter2,girdhar2024factorizing,kondratyuk2023videopoet}. More recently, diffusion transformers and MM-DiT architectures have shown stronger scalability in image and video generation~\cite{peebles2023scalable,esser2024scaling,ma2024latte,yang2024cogvideox,zheng2024open,kong2024hunyuanvideo,wan2025wan,polyak2024movie}. Our work builds on pre-trained video diffusion generators and improves multi-prompt temporal coherence without updating model parameters.

\paragraph{Long and multi-prompt video generation.}
Long video generation requires maintaining visual fidelity, temporal consistency, and semantic progression over extended durations. Training-based methods introduce dedicated long-horizon architectures, datasets, or autoregressive generation strategies~\cite{villegas2022phenaki,henschel2025streamingt2v,yan2025long,zhang2025frame,huang2025self}. In contrast, training-free methods reuse short-video generators through temporal co-denoising, noise rescheduling, sliding windows, queue-based generation, frequency-domain fusion, position mapping, or informative frame selection~\cite{wang2023gen,qiu2023freenoise,kim2024fifo,lu2024freelong,li2025longdiff}. Multi-prompt generation further requires prompt-specific semantic activation and smooth transitions across segment boundaries. Gen-L-Video, VideoDirectorGPT, VideoStudio, TALC, MEVG, Video-Infinity, and DiTCtrl explore multi-text, multi-scene, or multi-event video generation under different assumptions~\cite{wang2023gen,lin2023videodirectorgpt,long2024videostudio,bansal2024talc,oh2024mevg,tan2024video,cai2025ditctrl}. \textsc{FreqSync} focuses on training-free prompt-boundary coupling for coherent multi-prompt long videos.

\paragraph{Attention control and frequency-domain generation.}
Attention control enables training-free manipulation of diffusion generation by modifying attention behavior rather than model parameters~\cite{hertz2022prompt,cao2023masactrl,chefer2023attend,khachatryan2023text2video}. Recent studies also analyze or accelerate attention in video diffusion transformers, showing the importance of temporal correspondence and sparse spatiotemporal attention~\cite{nam2025emergent,zhang2025vsa,sun2025vorta}. In parallel, frequency-domain techniques have been used to improve initialization, temporal consistency, and long-video fidelity~\cite{wu2024freeinit,lu2024freelong}. Inspired by these works, \textsc{FreqSync} treats prompt boundaries as a frequency-sensitive coupling problem and jointly improves attention-level inheritance and prediction-level stitching.

\paragraph{Difference from DiTCtrl and frequency-domain long-video methods.}
\textsc{FreqSync} is distinct from both DiTCtrl~\cite{cai2025ditctrl} and existing frequency-domain long-video methods. DiTCtrl~\cite{cai2025ditctrl} improves multi-prompt transitions through cross-segment K/V sharing and latent blending, but such source reuse is largely frequency-agnostic and can over-preserve outdated layouts or semantics near prompt boundaries. In contrast, \textsc{FreqSync} keeps the target segment as the semantic anchor and introduces source information only through boundary-aligned frequency-selective routing: low- and mid-frequency components carry structure and motion continuity, while high-frequency details remain target-dominant. Frequency-domain long-video methods such as FreeLong~\cite{lu2024freelong} mainly enhance long-range temporal consistency through spectral feature blending, but they do not explicitly address the multi-prompt boundary trade-off between source continuity and target-prompt activation. \textsc{FreqSync} resolves this trade-off jointly at the attention level through FS-SCA and at the prediction level through RCSS, with residual clipping to prevent excessive boundary corrections.

\section{Method}
\label{sec:method}

\begin{figure*}[t]
    \centering
    \includegraphics[width=\textwidth]{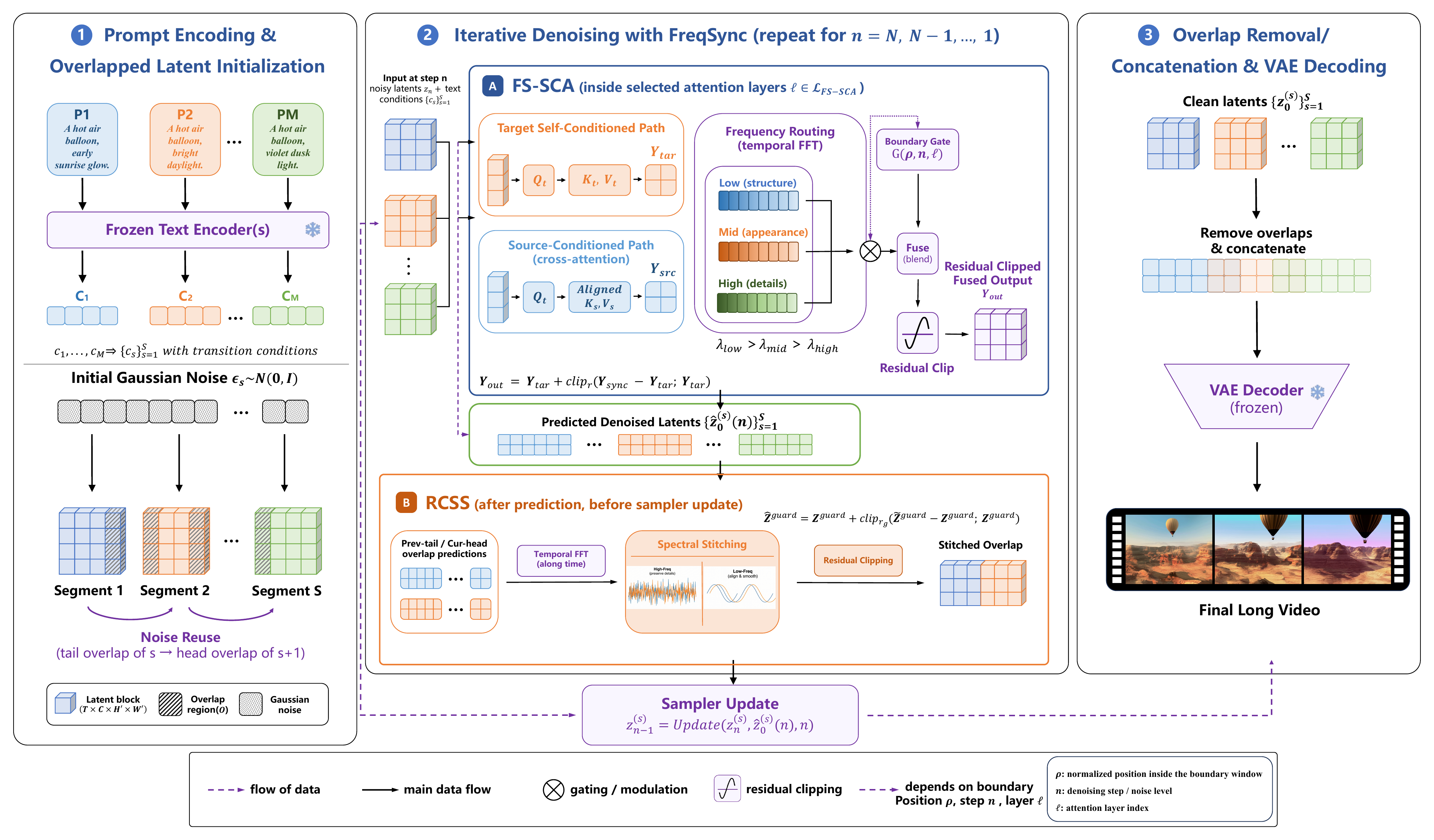}
    \vspace{-1.0em}
    \caption{Overview of \textsc{FreqSync}. Given a prompt sequence, \textsc{FreqSync} first encodes frozen text conditions and initializes overlapped latent segments with noise reuse. During iterative denoising, FS-SCA performs frequency-synchronized source-conditioned attention inside selected attention layers, while RCSS applies residual-clipped spectral stitching to denoised overlap predictions before the sampler update. After denoising, overlapped regions are removed and the clean latent segments are concatenated for VAE decoding, producing the final long video.}
    
    \label{fig:method_overview}
    \vspace{-1.0em}
\end{figure*}

We study training-free multi-prompt long video generation with a pre-trained video diffusion generator. Given an ordered prompt sequence, \textsc{FreqSync} synthesizes a coherent long video through overlapped latent segments, as shown in Figure~\ref{fig:method_overview}. It first encodes prompt conditions and initializes neighboring segments with overlap noise reuse. During denoising, FS-SCA operates inside selected attention layers, while RCSS stitches denoised overlap predictions before the sampler update. After denoising, overlapped regions are removed, the cleaned latent segments are concatenated, and the frozen VAE decoder produces the final long video.

\subsection{Problem Formulation}
\label{subsec:problem_formulation}

Let $\mathcal{P}=\{p_i\}_{i=1}^{M}$ denote an ordered sequence of prompts. We construct the long video by decomposing generation into overlapped latent segments. Let $T$ denote the latent length of each segment, $O$ the overlap size between neighboring segments, $B$ the number of transition segments inserted between adjacent prompts, and $L$ the number of additional dwell segments assigned to middle prompts. The total number of segments and the resulting long-video length are:
\begin{equation}
S = M + B(M-1) + L(M-2),
\qquad
T_{\mathrm{long}} = T + (S-1)(T-O).
\label{eq:problem_formulation}
\end{equation}
The first segment contributes $T$ latent frames, and each subsequent segment contributes $T-O$ new frames due to overlap reuse.

To initialize neighboring segments consistently, we reuse the overlap noise:
\begin{equation}
\boldsymbol{\epsilon}_{s+1}[1{:}O]
=
\boldsymbol{\epsilon}_{s}[T-O+1{:}T],
\label{eq:noise_reuse}
\end{equation}
where $\boldsymbol{\epsilon}_{s}$ denotes the initial latent noise of the $s$-th segment. For each adjacent prompt pair $(p_i,p_{i+1})$, we insert transition segments. The cross-attention condition for the $b$-th transition segment is interpolated by a cosine schedule:
\begin{equation}
\omega_{i,b}
=
\frac{1-\cos\!\left(\pi\frac{b+1}{B+1}\right)}{2},
\qquad
\mathbf{c}^{\mathrm{cross}}_{i,b}
=
(1-\omega_{i,b})\mathbf{c}^{\mathrm{cross}}_{i}
+
\omega_{i,b}\mathbf{c}^{\mathrm{cross}}_{i+1}.
\label{eq:transition_condition}
\end{equation}
Following the practical design of multi-prompt sampling, non-cross-attention branches are inherited from the dominant prompt rather than fully interpolated, which avoids off-manifold mixed conditions.

The key question is how neighboring segments should be coupled at a prompt boundary. Isolated generation causes discontinuities, while full source sharing or coarse blending may preserve outdated semantics. \textsc{FreqSync} resolves this tension in the temporal frequency domain: low- and mid-frequency source components carry structure and motion continuity, whereas high-frequency target components preserve prompt-specific details.

\subsection{Frequency-Synchronized Source-Conditioned Attention (FS-SCA)}
\label{subsec:source_conditioned_attention}

FS-SCA addresses the attention-level source-conditioning part of boundary coupling. Consider two neighboring segments, where the source segment corresponds to the previous prompt and the target segment corresponds to the current prompt. In an attention layer of the pre-trained video diffusion generator, let $(\mathbf{Q}_t,\mathbf{K}_t,\mathbf{V}_t)$ denote the query, key, and value features of the target segment, and let $(\mathbf{K}_s,\mathbf{V}_s)$ denote the key and value features of the source segment. Rather than directly replacing target features with source features, \textsc{FreqSync} computes two attention paths:

\vspace{-1.0em}
\begin{equation}
\mathbf{Y}^{\mathrm{tar}}
=
\operatorname{Attn}(\mathbf{Q}_t,\mathbf{K}_t,\mathbf{V}_t),
\qquad
\mathbf{Y}^{\mathrm{src}}
=
\operatorname{Attn}(\mathbf{Q}_t,\mathbf{K}^{\rightarrow}_{s},\mathbf{V}^{\rightarrow}_{s}).
\label{eq:dual_attention}
\end{equation}
The first path is the target self-conditioned path, which preserves prompt-specific semantics. The second path is the source-conditioned path, where $\mathbf{K}^{\rightarrow}_{s}$ and $\mathbf{V}^{\rightarrow}_{s}$ are source features aligned to the target boundary. Specifically, the tail overlap of the source segment is aligned to the head overlap of the target segment, so that the same global boundary frames are compared and fused. For the first few non-overlap guard frames, the final source-overlap representation is used as a structural reference. This tail-to-head alignment is important because local frame indices in neighboring segments do not correspond to the same global temporal positions.

We reshape the video feature part of $\mathbf{Y}^{\mathrm{tar}}$ and $\mathbf{Y}^{\mathrm{src}}$ into $T\times N\times d$ tensors, where $T$ is the compressed temporal length and $N$ is the number of spatial tokens. We then decompose each output into temporal low-, mid-, and high-frequency bands:
\begin{equation}
\mathbf{Y}^{b}
=
\mathcal{F}^{-1}\!\left(\mathbf{M}_{b}\odot\mathcal{F}(\mathbf{Y})\right),
\qquad
b\in\{\mathrm{low},\mathrm{mid},\mathrm{high}\},
\label{eq:freq_decomposition}
\end{equation}
where $\mathcal{F}$ denotes the temporal Fourier transform, $\mathbf{M}_{b}$ denotes a band mask, and $\odot$ denotes element-wise multiplication. This decomposition reflects the different roles of temporal frequency bands in boundary coupling: low-frequency components encode global layout, background, illumination, and coarse motion; mid-frequency components encode intermediate structure and local appearance; high-frequency components encode fine details and prompt-specific textures.

The fused video attention output is defined as:

\begin{equation}
\mathbf{Y}^{\mathrm{sync}}
=
\mathbf{Y}^{\mathrm{tar}}
+
G(\rho,n,\ell)
\sum_{b\in\{\mathrm{low},\mathrm{mid},\mathrm{high}\}}
\lambda_b
\left(
\mathbf{Y}^{\mathrm{src},b}
-
\mathbf{Y}^{\mathrm{tar},b}
\right).
\label{eq:spectral_fusion}
\end{equation}

Here, $G(\rho,n,\ell)$ controls where and when source-conditioned routing is active, while $\lambda_b$ controls how strongly each temporal frequency band inherits source-side continuity cues.

Accordingly, we choose the routing strengths to satisfy $\lambda_{\mathrm{low}}>\lambda_{\mathrm{mid}}>\lambda_{\mathrm{high}}$. In practice, $\lambda_{\mathrm{low}}$ is close to one near the boundary, $\lambda_{\mathrm{mid}}$ is moderate, and $\lambda_{\mathrm{high}}$ remains small. This design gives the target segment more source information where continuity matters, while preventing source high-frequency details from overwriting target prompt semantics.

To make the routing boundary-aware, we further modulate the routing strengths with a temporal boundary schedule. Let $\rho\in[0,1]$ denote the normalized temporal position inside the boundary window, and let $g(n,\ell)$ denote a smooth denoising-step and layer gate. The boundary gate is:
\begin{equation}
G(\rho,n,\ell)
=
g(n,\ell)\left[\cos\left(\frac{\pi\rho}{2}\right)\right]^{2\gamma},
\label{eq:boundary_gate}
\end{equation}
where $\gamma$ controls the decay from the source side to the target side. As a result, the gate is strongest near the beginning of the target segment and gradually decays as the target prompt takes over.

\paragraph{Residual clipping for stable source conditioning.}
While source-conditioned attention is essential for continuity, unconstrained correction may destabilize generation or over-constrain the target prompt. We therefore clip the frequency-fused correction relative to the target self-attention output. Let $\Delta=\mathbf{Y}^{\mathrm{sync}}-\mathbf{Y}^{\mathrm{tar}}$. The final output is:

\vspace{-1.0em}
\begin{equation}
\widehat{\mathbf{Y}}
=
\mathbf{Y}^{\mathrm{tar}}
+
\operatorname{clip}_{r}\!\left(\Delta;\mathbf{Y}^{\mathrm{tar}}\right),
\label{eq:residual_clipping}
\end{equation}
where $\operatorname{clip}_{r}$ limits the residual's frame-wise energy to a fixed ratio of the target output, preserving source-conditioned cues without excessive attention-output shifts.

\subsection{Residual-Clipped Spectral Stitching (RCSS)}
\label{subsec:spectral_stitching}

RCSS addresses the prediction-level stitching part of boundary coupling. Frequency-Synchronized Source-Conditioned Attention improves semantic and structural handover inside the denoising network, but the final long video also depends on how neighboring denoised segment predictions are stitched. Directly modifying the noisy solver state can push samples away from the model distribution and may cause severe artifacts. Coarse overlap blending, on the other hand, may suppress target details or propagate source-side residue. Therefore, \textsc{FreqSync} operates only on denoised overlap predictions and applies frequency-aware fusion with residual clipping to all post-overlap corrections.

Let $\mathbf{Z}^{\mathrm{prev}}\in\mathbb{R}^{O\times C\times H\times W}$ denote the tail overlap of the previous segment, and let $\mathbf{Z}^{\mathrm{cur}}\in\mathbb{R}^{O\times C\times H\times W}$ denote the head overlap of the current segment. We decompose each overlap into temporal bands as in Eq.~\eqref{eq:freq_decomposition}. The fused overlap is:
\begin{align}
\mathbf{Z}^{\mathrm{fuse}}
&=
\alpha_{\mathrm{low}}\mathbf{Z}^{\mathrm{prev,low}}
+(1-\alpha_{\mathrm{low}})\mathbf{Z}^{\mathrm{cur,low}}
\nonumber\\
&\quad+
\alpha_{\mathrm{high}}\mathbf{Z}^{\mathrm{prev,high}}
+(1-\alpha_{\mathrm{high}})\mathbf{Z}^{\mathrm{cur,high}},
\label{eq:denoised_fusion}
\end{align}
where $\alpha_{\mathrm{low}}$ follows a raised-cosine transition and $\alpha_{\mathrm{high}}$ decays faster to avoid ghosting. We additionally apply a light low-frequency temporal smoothing pass inside the overlap to reduce boundary-local flicker while preserving high-frequency target details.

To handle the non-overlap guard frames immediately after the boundary, we propagate the low-frequency trend from the fused overlap. Let $\mathbf{Z}^{\mathrm{guard}}$ denote the first $G$ non-overlap frames of the target segment. We compute a low-frequency memory $\mathbf{M}^{\mathrm{low}}$ from $\mathbf{Z}^{\mathrm{fuse}}$ and form a proposal:
\begin{equation}
\widetilde{\mathbf{Z}}^{\mathrm{guard}}
=
\mathbf{Z}^{\mathrm{guard,high}}
+
(1-\eta)\mathbf{Z}^{\mathrm{guard,low}}
+
\eta\mathbf{M}^{\mathrm{low}},
\label{eq:guard_memory}
\end{equation}
where $\eta$ is a decaying guard strength. The final guard correction is residual-clipped:
\begin{equation}
\widehat{\mathbf{Z}}^{\mathrm{guard}}
=
\mathbf{Z}^{\mathrm{guard}}
+
\operatorname{clip}_{r_g}\!\left(
\widetilde{\mathbf{Z}}^{\mathrm{guard}}-\mathbf{Z}^{\mathrm{guard}};
\mathbf{Z}^{\mathrm{guard}}
\right).
\label{eq:guard_clipping}
\end{equation}
This design is intentionally conservative: it synchronizes low-frequency continuation near boundaries, but avoids direct modification of the solver state and limits the magnitude of each correction.

\subsection{Inference Procedure}
\label{subsec:inference}

During inference, \textsc{FreqSync} expands the prompt sequence into overlapped latent segments with transition conditions and performs iterative denoising over neighboring segment pairs. The complete inference procedure is summarized in Appendix~\ref{app:inference_pseudocode}. In scheduled layers and timesteps, FS-SCA evaluates the target segment through both target self-conditioned and source-conditioned attention paths, then applies frequency-selective routing with residual clipping so that low- and mid-frequency continuity cues are inherited while high-frequency target details remain dominant. After each denoised prediction is obtained, RCSS fuses the boundary overlap and propagates conservative low-frequency continuation to boundary-adjacent guard frames. All corrections are applied to attention outputs or denoised predictions rather than directly to the noisy solver state, keeping the target branch as the semantic anchor while using the source segment only as a boundary-continuity reference.

\begin{table*}[t]
    \centering
    \small
    \setlength{\tabcolsep}{5.0pt}
    \renewcommand{\arraystretch}{1.12}
    \caption{Quantitative comparison. Results are averaged over three seeds on the 130-case multi-prompt benchmark. \textsc{FreqSync} delivers the strongest overall performance, achieving the best scores on CSCV, Text-Image Similarity, T2V Alignment, and Physical Consistency, while remaining highly competitive on Motion Smoothness and Visual Quality with only marginal gaps to the best results. Best scores are in \textbf{bold}; second-best scores are \underline{underlined}.}

    \label{tab:main_quantitative_results}
    \resizebox{\textwidth}{!}{
    \begin{tabular}{lcccccc}
        \toprule
        Method 
        & \makecell{CSCV\\(\%)$\uparrow$} 
        & \makecell{Motion\\Smooth. (\%)$\uparrow$} 
        & \makecell{Text-Image\\Similarity$\uparrow$} 
        & \makecell{Visual\\Quality$\uparrow$} 
        & \makecell{T2V\\Align.$\uparrow$} 
        & \makecell{Physical\\Consist.$\uparrow$} \\
        \midrule
        MEVG~\cite{oh2024mevg} 
        & 67.22 & 95.66 & 29.56 & 1.970 & 2.773 & 2.280 \\
        Video-Infinity~\cite{tan2024video} 
        & 83.62 & 96.45 & 29.72 & \textbf{3.247} & 3.422 & 3.210 \\
        DiTCtrl~\cite{cai2025ditctrl} 
        & \underline{86.11} & \textbf{98.44} & \underline{30.33} 
        & 3.147 & \underline{3.758} & \underline{3.228} \\
        \midrule[0.8pt]
        \rowcolor{gray!12}
        \textsc{FreqSync} (Ours) 
        & \textbf{86.43} & \underline{98.25} & \textbf{30.53} 
        & \underline{3.214} & \textbf{3.849} & \textbf{3.240} \\
        \bottomrule
    \end{tabular}}
    \vspace{-1.0em}
\end{table*}
\section{Experiments}
\label{sec:experiments}

\subsection{Experimental Setup}
\label{subsec:experimental_setup}

\paragraph{Implementation details.}

We implement \textsc{FreqSync} on the pre-trained CogVideoX-2B backbone~\cite{yang2024cogvideox} in a fully training-free manner. Videos are generated at $480\times720$ resolution and 16 FPS. We use latent segments of 13 frames, an overlap size of 9, and 2 transition blocks between adjacent prompts. FS-SCA is applied to denoising steps $[2,31)$ and transformer layers $[22,30)$. Its temporal Fourier bands are split by normalized low- and mid-frequency cutoff ratios of 0.30 and 0.52, which define the low/mid/high bands used for frequency routing. Residual clipping is used for both FS-SCA attention fusion and RCSS guard correction. All experiments are performed on NVIDIA A800 GPUs.

\paragraph{Benchmark, baselines, and metrics.}
Following the MPVBench setting~\cite{cai2025ditctrl}, we construct a 130-case multi-prompt benchmark covering ten transition modes: background, subject, camera, style, lighting, location, speed, emotion, clothing, and action. Each method is evaluated over three random seeds. We compare with MEVG~\cite{oh2024mevg}, Video-Infinity~\cite{tan2024video}, and DiTCtrl~\cite{cai2025ditctrl}; qualitative comparisons additionally include a commercial model, Kling-2.6~\cite{kling2024official}. We report CSCV for transition smoothness, VBench Motion Smoothness~\cite{huang2024vbench}, CLIP-based Text-Image Similarity~\cite{radford2021learning}, and VideoScore2 dimensions including Visual Quality, T2V Alignment, and Physical Consistency~\cite{he2025videoscore2}.

\subsection{Quantitative Results}
\label{subsec:quantitative_results}

Table~\ref{tab:main_quantitative_results} reports the main quantitative comparison across transition smoothness, temporal motion quality, prompt alignment, and perceptual video quality. \textsc{FreqSync} achieves the strongest overall performance, ranking first on four out of six metrics and second on the remaining two with only marginal gaps to the best scores. Compared with MEVG and Video-Infinity~\cite{oh2024mevg,tan2024video}, our method substantially improves CSCV and motion-related quality, indicating more stable cross-segment evolution. Compared with DiTCtrl~\cite{cai2025ditctrl}, \textsc{FreqSync} further improves CSCV from 86.11\% to 86.43\%, while maintaining a very close Motion Smoothness score of 98.25\% versus 98.44\%. More importantly, \textsc{FreqSync} achieves the highest Text-Image Similarity and T2V Alignment scores, suggesting that frequency-selective boundary coupling improves target-prompt activation without over-preserving outdated source semantics. Although Video-Infinity obtains the highest Visual Quality score, \textsc{FreqSync} remains close and achieves stronger transition smoothness, text alignment, and physical consistency. These results demonstrate that \textsc{FreqSync} provides the best overall boundary-coupling trade-off. Further per-seed results, significance tests, cross-backbone evaluations, and efficiency analyses are provided in Appendices~\ref{app:additional_quantitative}, \ref{app:statistical_significance}, \ref{app:cross_backbone}, and~\ref{app:efficiency_analysis}.

\begin{figure*}[t]
    \centering
    \includegraphics[width=\linewidth]{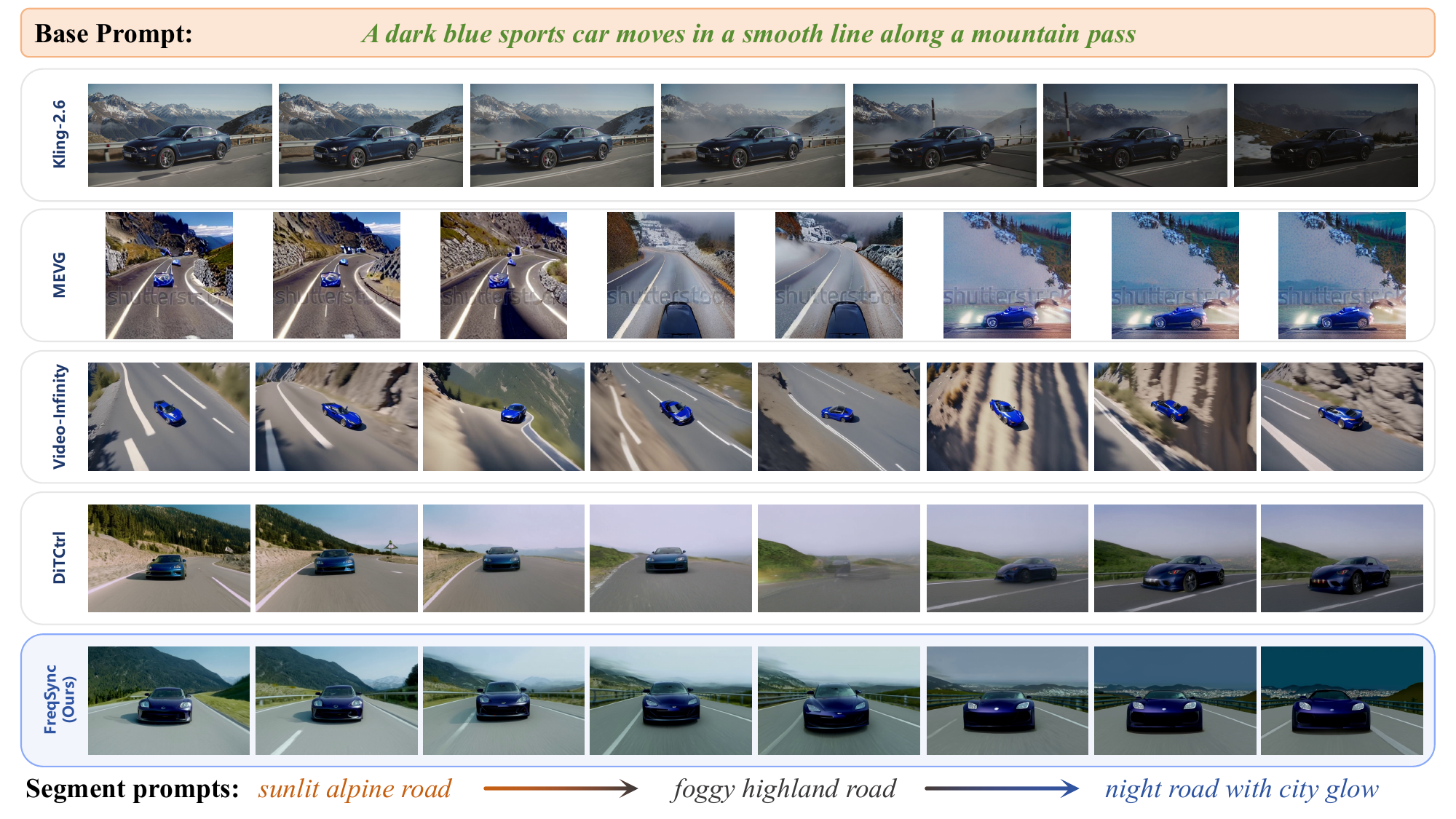}
    \vspace{-1.0em}
    \caption{Qualitative comparison. On a three-prompt driving sequence, baselines suffer from viewpoint drift, weak prompt transition, ghosting, or duplicated traces. \textsc{FreqSync} preserves vehicle identity and motion while producing smoother semantic transitions.}
    
    \label{fig:qualitative_results}

    \vspace{-1.0em}
\end{figure*}

\subsection{Qualitative Results}
\label{subsec:qualitative_results}

Figure~\ref{fig:qualitative_results} presents a representative qualitative comparison on a challenging three-prompt sequence, where a dark blue sports car should move from a sunlit alpine road to a foggy highland road and finally to a night road with city glow, while preserving its identity and motion trajectory. MEVG~\cite{oh2024mevg} shows severe cross-segment discontinuities: the camera viewpoint, road layout, and vehicle scale change abruptly, and repeated or inconsistent subjects appear. Video-Infinity~\cite{tan2024video} produces weak dynamics and unstable visual details, with nearly static motion, flickering backgrounds, and ghosting artifacts. DiTCtrl~\cite{cai2025ditctrl} better maintains vehicle identity and global continuity, but its full-frequency source inheritance still introduces boundary artifacts during the foggy-to-night transition, including ghosting, duplicated visual traces, and an abrupt semantic switch. Kling-2.6~\cite{kling2024official} generates realistic frames but shows weaker prompt following: the foggy road and final city-glow semantics are insufficiently expressed. In contrast, \textsc{FreqSync} preserves a stable car identity, viewpoint, and forward motion while progressively changing the background; the fog and night illumination emerge smoothly without duplicated subjects or strong ghosting. This comparison supports our frequency-synchronized boundary coupling design: low- and mid-frequency synchronization maintains structure and motion, while target-dominant high-frequency components preserve prompt-specific details. More qualitative results and the single-prompt long-video application are provided in Appendix~\ref{app:additional_qualitative}.

\subsection{Diagnostic Analysis of Boundary Coupling}
\label{subsec:diagnostic_analysis}

\begin{wrapfigure}{l}{0.58\linewidth}
    \vspace{-0.8em}
    \centering
    \includegraphics[width=\linewidth]{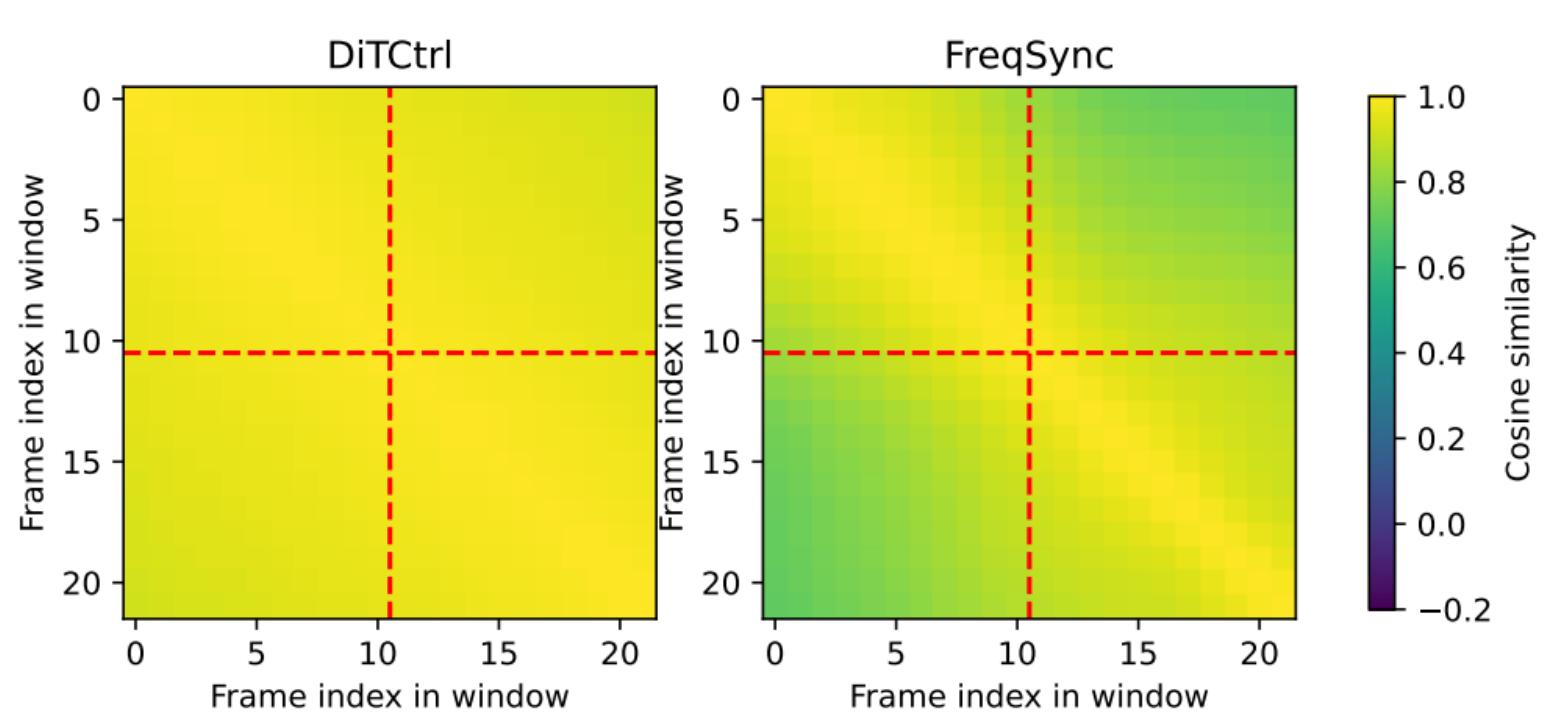}
    \vspace{-1.2em}
    \caption{Temporal self-similarity around prompt boundaries. \textsc{FreqSync} preserves local diagonal continuity while reducing excessive cross-boundary similarity, supporting frequency-selective boundary coupling.}
    \label{fig:temporal_similarity_map}
    \vspace{-0.6em}
\end{wrapfigure}

We further analyze the frequency-agnostic boundary coupling problem identified in Sec.~\ref{sec:introduction}. Figure~\ref{fig:temporal_similarity_map} visualizes temporal self-similarity matrices around a prompt boundary. DiTCtrl~\cite{cai2025ditctrl} exhibits uniformly high cross-boundary similarity, indicating strong source-segment retention. While this helps smooth transitions, it may also preserve outdated source layouts and correlate with duplicated-subject artifacts, as observed in Figure~\ref{fig:diagnostic_overconditioning}. In contrast, \textsc{FreqSync} maintains strong local temporal continuity along the diagonal while reducing excessive cross-boundary similarity, suggesting a better balance between source continuity and target-prompt activation. This analysis supports our motivation: smoothness alone is not sufficient for multi-prompt generation. A successful method should selectively transfer continuity cues, preserve emerging target details, and stitch adjacent denoised predictions without high-energy boundary corrections. \textsc{FreqSync} enforces this balance through FS-SCA and RCSS, which jointly improve attention-level source conditioning and prediction-level boundary stitching. Further temporal-frequency analysis is provided in Appendix~\ref{app:additional_diagnostic}.

\begin{figure*}[t]
    \centering
    \includegraphics[width=\linewidth]{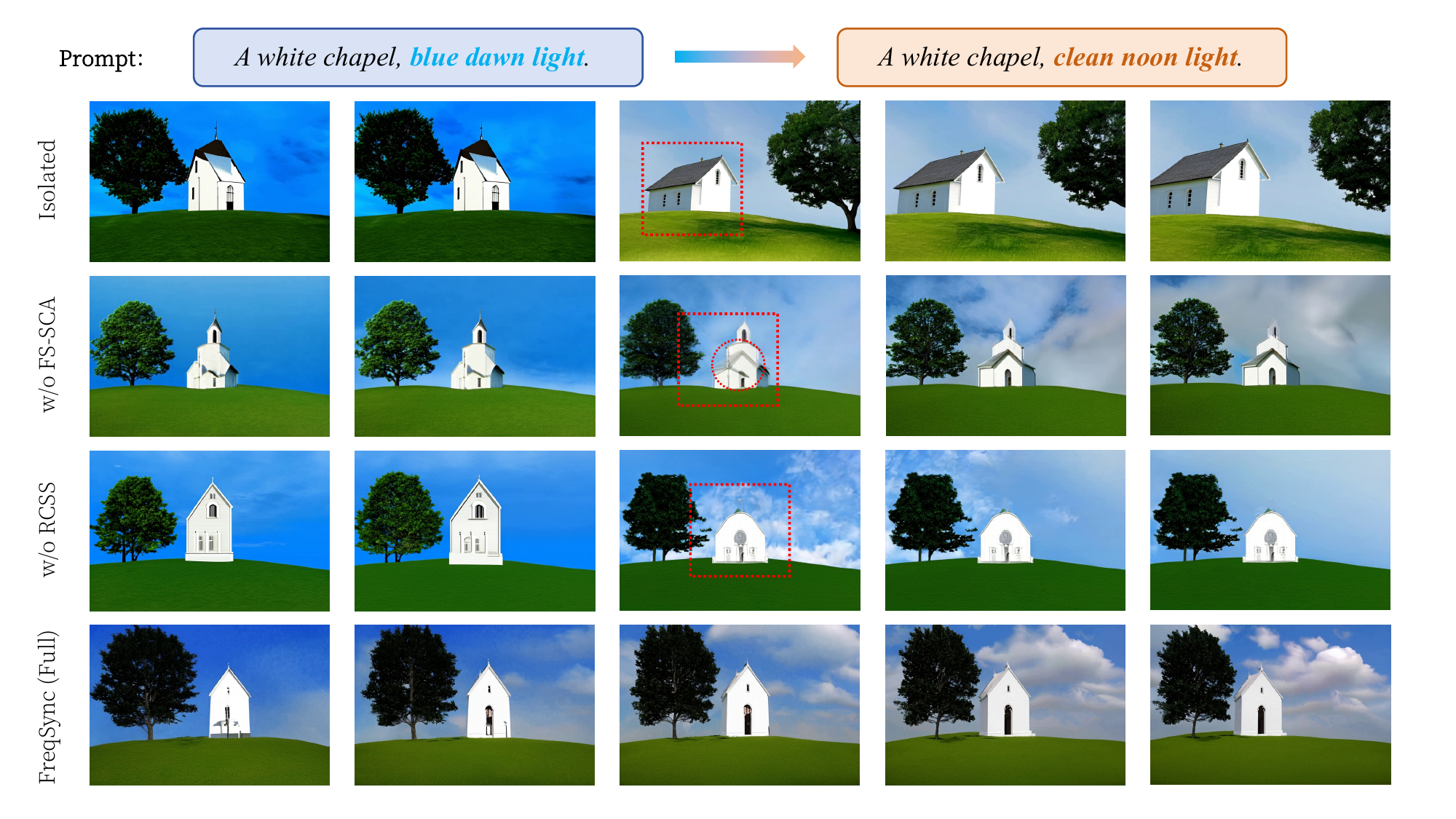}
    \vspace{-1.0em}
    \caption{Visual ablation of \textsc{FreqSync}.
    Isolated generation produces a hard cut. RCSS alone smooths the overlap region but lacks attention-level semantic handover. FS-SCA alone preserves the global layout but changes the chapel structure. Full \textsc{FreqSync} combines both modules for smooth transitions and stable identity.}

    \label{fig:ablation_results}

    \vspace{-0.6em}
\end{figure*}

\begin{table}[t]
    \caption{Ablation study on a 40-case subset.
    Ablations are conducted on a representative 40-case subset covering all transition modes under a single random seed. CSCV is the primary boundary-transition metric. Segment-level Text-Image Similarity is reported as a prompt-fidelity reference and is not boundary-aware. Best results are in \textbf{bold} and second-best results are \underline{underlined}.}

    \label{tab:ablation_results}
    \centering
    \small
    \setlength{\tabcolsep}{5pt}
    \begin{tabular}{lcccc}
        \toprule
        Variant & FS-SCA & RCSS & CSCV (\%)$\uparrow$ & Text-Image Sim.$\uparrow$ \\
        \midrule
        Isolated & $\times$ & $\times$ & 76.96 & \textbf{30.89} \\
        w/o FS-SCA (RCSS only) & $\times$ & $\checkmark$ & \underline{85.73} & 30.62 \\
        w/o RCSS (FS-SCA only) & $\checkmark$ & $\times$ & 82.08 & 30.56 \\
        Full \textsc{FreqSync} & $\checkmark$ & $\checkmark$ & \textbf{89.14} & \underline{30.71} \\
        \bottomrule
    \end{tabular}
    \vspace{-1.5em}
\end{table}

\vspace{-0.3em}
\subsection{Ablation Studies}
\label{subsec:ablation_studies}

We ablate the two core components of \textsc{FreqSync}: FS-SCA and RCSS. Ablation results are evaluated on a 40-case subset under a single random seed. As shown in Table~\ref{tab:ablation_results} and Figure~\ref{fig:ablation_results}, isolated generation produces abrupt prompt-boundary changes and obtains the lowest CSCV score. Using RCSS alone improves CSCV from 76.96\% to 85.73\%, showing that prediction-level spectral stitching can smooth the overlap region, but the lack of FS-SCA still causes slight changes in object appearance and structural handover. Using FS-SCA alone improves CSCV to 82.08\%, indicating that attention-level source conditioning helps preserve layout and transition continuity, although the chapel structure still changes without prediction-level stitching. The full \textsc{FreqSync} achieves the best CSCV score of 89.14\% and produces the most coherent transition: the global layout, chapel position, and object appearance remain stable while the illumination follows the target prompt. Segment-level Text-Image Similarity serves as a prompt-fidelity reference; although isolated generation obtains a slightly higher score because it is not boundary-aware, full \textsc{FreqSync} achieves the best prompt alignment among coupled variants with much smoother transitions. Additional frequency-routing ablations and hyperparameter sensitivity analyses are provided in Appendices~\ref{app:routing_ablation} and~\ref{app:sensitivity_analysis}, respectively.

\subsection{Human Evaluation}
\label{subsec:human_evaluation}

To complement automatic metrics, we conduct a blind human evaluation with 30 participants on 40 anonymized cases covering all ten transition modes and three random seeds. Four methods are evaluated on a five-point scale across five dimensions: target semantic correctness, source leakage and artifacts, transition smoothness, visual quality, and overall preference. Higher scores indicate better performance, with higher scores for source leakage and artifacts reflecting fewer unwanted artifacts. As shown in Table~\ref{tab:human_study}, \textsc{FreqSync} achieves the highest mean scores across all five dimensions, with particularly strong improvements in target semantic correctness, artifact suppression, and transition smoothness. These results further demonstrate that \textsc{FreqSync} effectively improves temporal continuity while preserving newly introduced prompt semantics, consistent with the quantitative and qualitative evaluations. Additional protocol details are provided in Appendix~\ref{app:human_protocol}.

\begin{table*}[t]
    \centering
    \small
    \setlength{\tabcolsep}{5pt}
    \renewcommand{\arraystretch}{1.10}
    \caption{Blind human evaluation on 40 anonymized cases with 30 participants. Mean five-point scores are reported, where higher is better in every column. For source leakage and artifacts, higher scores mean fewer artifacts. Best scores are in \textbf{bold}.}
    \vspace{0.5em}
    \label{tab:human_study}
    \resizebox{\textwidth}{!}{%
    \begin{tabular}{lccccc}
    \toprule
    Method
    & \makecell{Target semantic\\correctness$\uparrow$}
    & \makecell{Source leakage\\and artifacts$\uparrow$}
    & \makecell{Transition\\smoothness$\uparrow$}
    & \makecell{Visual\\quality$\uparrow$}
    & \makecell{Overall\\preference$\uparrow$} \\
    \midrule
    MEVG~\cite{oh2024mevg}
    & 1.67 & 1.53 & 1.70 & 1.83 & 1.63 \\
    Video-Infinity~\cite{tan2024video}
    & 2.70 & 2.57 & 2.83 & 2.97 & 2.70 \\
    DiTCtrl~\cite{cai2025ditctrl}
    & 3.33 & 3.07 & 2.80 & 2.90 & 3.17 \\
    \midrule
    \textsc{FreqSync} (Ours)
    & \textbf{4.30} & \textbf{4.23}
    & \textbf{4.27} & \textbf{4.07}
    & \textbf{4.30} \\
    \bottomrule
    \end{tabular}}
    \vspace{-1.5em}
\end{table*}

\section{Conclusion}
\label{sec:conclusion}

We present \textsc{FreqSync}, a training-free framework for multi-prompt long video generation based on frequency-synchronized boundary coupling. Starting from the observation that weak boundary coupling causes abrupt transitions whereas overly strong or frequency-agnostic coupling may preserve outdated source semantics and introduce ghosting or duplicated subjects, \textsc{FreqSync} decomposes prompt transitions into attention-level source conditioning and prediction-level boundary stitching. FS-SCA transfers low- and mid-frequency continuity cues while preserving target-dominant details, and RCSS aligns denoised overlap predictions under residual constraints for stable stitching. Extensive quantitative, qualitative, diagnostic, and ablation results demonstrate that \textsc{FreqSync} improves transition smoothness, prompt alignment, and overall video quality without additional training, providing an effective solution for coherent multi-prompt long video generation.

\begin{ack}
This work was supported by the National Natural Science Foundation of China (No. 62576109), the Scientific and Technological Innovation Action Plan of the Shanghai Science and Technology Committee (No. 25511104402), and the AI for Science Program of the Shanghai Municipal Commission of Economy and Informatization (No. 2025-GZL-RGZN-BTBX-02006). The authors declare no competing interests.

\end{ack}

\clearpage

\bibliographystyle{plainnat}
\bibliography{references}

\clearpage


\appendix

\section*{Appendix}
\begin{itemize}
    \item In Appendix~\ref{app:algorithmic_details}, we provide additional algorithmic and implementation details of \textsc{FreqSync}, including the inference pseudocode, the placement of RCSS before the sampler update, the supplementary code organization, and the configuration hyperparameters.

    \item In Appendix~\ref{app:bounded_residual}, we provide a stability analysis of the residual clipping mechanism used in FS-SCA and RCSS, showing that each introduced correction is locally bounded relative to a reference representation.

    \item In Appendix~\ref{app:evaluation_details}, we provide additional evaluation details, including case-level aggregation, prompt-level text alignment, and the random seed setting used for quantitative evaluation.

    \item In Appendix~\ref{app:additional_quantitative}, we provide additional quantitative results, including per-seed results for all compared methods under three random seeds.

    \item In Appendix~\ref{app:statistical_significance}, we report paired significance tests on 100 balanced cases, including bootstrap confidence intervals and Holm-adjusted tests.

    \item In Appendix~\ref{app:cross_backbone}, we evaluate Wan2.1 and VideoCrafter2 to examine cross-architecture transfer and its limitations.

    \item In Appendix~\ref{app:routing_ablation}, we compare low/mid/high source-routing controls, including reverse and uniform routing.

    \item In Appendix~\ref{app:sensitivity_analysis}, we report OFAT and cutoff--layer interaction tests.

    \item In Appendix~\ref{app:efficiency_analysis}, we report sampling time, GPU memory, and counted FLOPs.

    \item In Appendix~\ref{app:additional_diagnostic}, we provide additional diagnostic visualizations that analyze temporal-frequency energy around prompt boundaries.

    \item In Appendix~\ref{app:additional_qualitative}, we provide additional qualitative visualizations, including more generated examples of \textsc{FreqSync}, a single-prompt long-video generation application, and additional qualitative comparisons with Kling-2.6, MEVG, Video-Infinity, and DiTCtrl~\cite{kling2024official,oh2024mevg,tan2024video,cai2025ditctrl}.

    \item In Appendix~\ref{app:broader_impacts}, we discuss the broader impacts of \textsc{FreqSync}, including potential positive applications and possible misuse risks of coherent long-video generation.

    \item In Appendix~\ref{app:licenses}, we summarize the existing models, baselines, benchmarks, and evaluation tools used in this work, together with their corresponding licenses or usage terms.

    \item In Appendix~\ref{app:limitations}, we discuss the limitations of \textsc{FreqSync}, including dependence on the underlying pre-trained generator, challenges under large semantic changes, additional inference overhead, and inherited model risks.
\end{itemize}

\section{Algorithmic and Implementation Details}
\label{app:algorithmic_details}

\subsection{Inference Pseudocode}
\label{app:inference_pseudocode}

Algorithm~\ref{alg:freqsync} summarizes the inference procedure of \textsc{FreqSync}. Unlike the method overview in the main paper, this pseudocode clarifies the order in which prompt expansion, overlapped latent initialization, FS-SCA, RCSS, sampler update, and final latent concatenation are applied.

\begin{algorithm}[!t]
\caption{\textsc{FreqSync} Inference}
\label{alg:freqsync}
\small
\begin{algorithmic}[1]
\REQUIRE Prompt sequence $\mathcal{P}=\{p_i\}_{i=1}^{M}$, frozen video diffusion generator $\epsilon_\theta$, segment length $T$, overlap length $O$, transition blocks $B$, dwell segments $L$
\STATE Compute $S$ and $T_{\mathrm{long}}$ by Eq.~\eqref{eq:problem_formulation}.
\STATE Build transition conditions $\mathbf{c}^{\mathrm{cross}}_{i,b}$ by Eq.~\eqref{eq:transition_condition}.
\STATE Initialize latent segments $\{\mathbf{x}^{(N)}_s\}_{s=1}^{S}$ with overlap noise reuse by Eq.~\eqref{eq:noise_reuse}.

\FOR{each denoising step $n=N,\ldots,1$}
    \FOR{each neighboring segment pair $(s,s+1)$}
        \STATE Run $\epsilon_\theta$ on the neighboring segments.
        \IF{$(n,\ell)$ is scheduled for FS-SCA}
            \STATE Compute $\mathbf{Y}^{\mathrm{tar}}$ and $\mathbf{Y}^{\mathrm{src}}$ by Eq.~\eqref{eq:dual_attention} with tail-to-head aligned source K/V.
            \STATE Decompose attention outputs into temporal bands by Eq.~\eqref{eq:freq_decomposition}.
            \STATE Obtain $\mathbf{Y}^{\mathrm{sync}}$ using the boundary gate in Eq.~\eqref{eq:boundary_gate} and the routing rule in Eq.~\eqref{eq:spectral_fusion}.
            \STATE Obtain the clipped attention output $\widehat{\mathbf{Y}}$ by Eq.~\eqref{eq:residual_clipping}.
        \ENDIF

        \STATE Obtain denoised predictions for the neighboring segments.
        \STATE Extract $\mathbf{Z}^{\mathrm{prev}}$ and $\mathbf{Z}^{\mathrm{cur}}$ from the boundary overlap.
        \STATE Fuse the denoised overlap $\mathbf{Z}^{\mathrm{fuse}}$ by Eq.~\eqref{eq:denoised_fusion}.
        \STATE Apply low-frequency guard continuation by Eq.~\eqref{eq:guard_memory}.
        \STATE Apply residual-clipped guard correction by Eq.~\eqref{eq:guard_clipping}.
        \STATE Update noisy latent states with the original diffusion sampler.
    \ENDFOR
\ENDFOR

\STATE Remove duplicated overlaps and concatenate the cleaned latent segments.
\STATE Decode the final latent sequence with the frozen VAE decoder.
\RETURN Generated multi-prompt long video.
\end{algorithmic}
\end{algorithm}

\subsection{Why RCSS Is Applied Before the Sampler Update}
\label{app:before_sampler_update}

RCSS is applied to denoised overlap predictions rather than directly to the noisy solver state. A diffusion sampler updates the noisy latent according to the current denoised prediction and the noise schedule. Directly modifying the noisy state may introduce uncontrolled perturbations that are further propagated by later denoising steps. In contrast, modifying denoised predictions allows boundary corrections to enter the update through the original sampler rule. Therefore, RCSS refines the model prediction at the boundary while keeping the solver dynamics unchanged. This design makes the boundary correction more conservative and reduces the risk of off-manifold latent updates.

\subsection{Supplementary Code and Configuration}
\label{app:supplementary_code}

We include partial source code, the corresponding configuration file, and representative prompt examples in the supplementary material. The code is organized into three files. \texttt{freqsync\_modules.py} contains the frequency-domain utilities shared by FS-SCA and RCSS, including temporal band decomposition, source-aligned K/V construction, spectral fusion, and residual clipping. \texttt{freqsync\_attention.py} contains the attention-layer integration of FS-SCA, where the target segment is processed through both target self-conditioned and source-conditioned attention paths. \texttt{freqsync\_sampler.py} contains the multi-prompt inference logic, including prompt-sequence expansion, overlap noise reuse, and RCSS before the sampler update. The file \texttt{config.yaml} lists the main sampling and frequency-routing hyperparameters used by \textsc{FreqSync}. We also include \texttt{sample\_prompts.json}, which provides representative prompt-sequence examples for each transition mode and illustrates the input format for multi-prompt generation. 


\subsection{Configuration Hyperparameters}
\label{app:configuration_hyperparameters}

Table~\ref{tab:freqsync_config_hyperparameters} summarizes the main hyperparameters included in \texttt{config.yaml}. Basic inference parameters determine how an input prompt sequence is decomposed into overlapped latent segments. FS-SCA parameters specify where attention-level source conditioning is applied and how source information is routed across temporal frequency bands. RCSS parameters control prediction-level spectral stitching around the overlap, including low-pass seam alignment, post-overlap guard synchronization, and residual-energy clipping.

\begin{table}[h]
\centering
\caption{Main hyperparameters in the supplementary configuration file. The table lists the core parameters included in \texttt{config.yaml}. FS-SCA controls frequency-selective attention-level source conditioning, while RCSS controls residual-constrained prediction-level boundary stitching.}

\label{tab:freqsync_config_hyperparameters}
\resizebox{\linewidth}{!}{
\begin{tabular}{llcl}
\toprule
Module & Hyperparameter & Value & Description \\
\midrule
Basic inference & latent frames per segment & 13 & Number of latent frames generated for each segment. \\
Basic inference & sampling FPS & 16 & Frame rate used when saving the generated video. \\
Basic inference & overlap size & 9 & Number of shared latent frames between adjacent segments. \\
Basic inference & transition blocks & 2 & Number of transition segments inserted between neighboring prompts. \\
Basic inference & seed & 42 & Default seed used by the example configuration. \\
\midrule
FS-SCA & active attention layers & $[22,30)$ & Transformer layers where source-conditioned attention is applied. \\
FS-SCA & active denoising steps & $[2,31)$ & Denoising steps where FS-SCA is enabled. \\
FS-SCA & low-frequency cutoff & 0.30 & Temporal cutoff used to extract stable low-frequency structure. \\
FS-SCA & mid-frequency cutoff & 0.52 & Temporal cutoff separating mid-frequency motion from high-frequency details. \\
FS-SCA & source low-frequency strength & 0.98 & Weight for transferring source-side low-frequency continuity cues. \\
FS-SCA & source mid-frequency strength & 0.68 & Weight for transferring source-side mid-frequency motion cues. \\
FS-SCA & source high-frequency strength & 0.035 & Small source weight retained in high-frequency bands to keep target details dominant. \\
FS-SCA & attention-output residual clip & 0.32 & Maximum relative residual energy allowed for attention-output correction. \\
\midrule
RCSS & boundary low-pass ratio & 0.23 & Low-pass cutoff used for spectral seam alignment in denoised predictions. \\
RCSS & boundary smoothing strength & 0.12 & Strength of temporal smoothing applied near the overlap. \\
RCSS & boundary smoothing steps & 32 & Number of early/middle denoising steps using boundary smoothing. \\
RCSS & guard frames & 4 & Number of post-overlap frames receiving low-frequency continuation. \\
RCSS & guard strength & 0.52 & Strength of the post-overlap low-frequency guard. \\
RCSS & guard residual clip & 0.18 & Maximum relative residual energy allowed for guard-frame correction. \\
\bottomrule
\end{tabular}}
\end{table}

The source routing strengths are intentionally larger for low- and mid-frequency bands, which mainly carry layout, trajectory, and coarse motion continuity across prompt boundaries. The high-frequency source strength is kept small so that fine target-prompt details remain dominant. Residual clipping is applied in both FS-SCA and RCSS to prevent overly strong boundary corrections and reduce semantic contamination or stitching artifacts.

\section{Bounded Residual Correction}
\label{app:bounded_residual}

This section provides a simple stability analysis for the residual clipping used in FS-SCA and RCSS. The purpose is not to prove global convergence of the diffusion sampler, but to show that each correction introduced by \textsc{FreqSync} is locally bounded relative to a reference representation.

\subsection{RMS-Norm Residual Bound}
\label{app:rms_bound}

Let $\mathbf{B}$ denote the base representation, $\widetilde{\mathbf{B}}$ denote the proposed corrected representation, and $\mathbf{R}$ denote the reference representation. Define the residual as
\begin{equation}
\Delta = \widetilde{\mathbf{B}}-\mathbf{B}.
\end{equation}
The residual-clipped correction used in \textsc{FreqSync} can be written as
\begin{equation}
\widehat{\mathbf{B}}
=
\mathbf{B}
+
s\Delta,
\qquad
s=
\min
\left(
1,
\frac{r\cdot \operatorname{RMS}(\mathbf{R})}
{\operatorname{RMS}(\Delta)+\epsilon}
\right),
\label{eq:appendix_residual_clipping}
\end{equation}
where $r$ is the residual clipping ratio and $\epsilon$ is a small constant for numerical stability. The resulting correction satisfies
\begin{equation}
\operatorname{RMS}
\left(
\widehat{\mathbf{B}}-\mathbf{B}
\right)
\le
r\cdot \operatorname{RMS}(\mathbf{R}).
\label{eq:appendix_residual_bound}
\end{equation}

\paragraph{Proof.}
Since $\widehat{\mathbf{B}}-\mathbf{B}=s\Delta$, we have
\begin{equation}
\operatorname{RMS}
\left(
\widehat{\mathbf{B}}-\mathbf{B}
\right)
=
s\operatorname{RMS}(\Delta).
\end{equation}
If $s=1$, then by the definition of $s$,
\begin{equation}
\operatorname{RMS}(\Delta)
\le
r\operatorname{RMS}(\mathbf{R}).
\end{equation}
If $s<1$, then
\begin{equation}
s\operatorname{RMS}(\Delta)
=
\frac{r\operatorname{RMS}(\mathbf{R})}
{\operatorname{RMS}(\Delta)+\epsilon}
\operatorname{RMS}(\Delta)
\le
r\operatorname{RMS}(\mathbf{R}).
\end{equation}
Thus, in both cases,
\begin{equation}
\operatorname{RMS}
\left(
\widehat{\mathbf{B}}-\mathbf{B}
\right)
\le
r\operatorname{RMS}(\mathbf{R}).
\end{equation}
\hfill $\square$

\paragraph{Implication for FS-SCA.}
For FS-SCA, $\mathbf{B}$ is the target self-conditioned attention output, $\widetilde{\mathbf{B}}$ is the frequency-fused source-conditioned proposal, and $\mathbf{R}$ is the target attention output. Therefore, the source-conditioned attention correction is bounded by a fixed ratio of the target-path activation energy. This prevents source-conditioned attention from fully overriding the target branch.

\paragraph{Implication for RCSS.}
For RCSS, $\mathbf{B}$ is the original denoised guard prediction, $\widetilde{\mathbf{B}}$ is the low-frequency continuation proposal, and $\mathbf{R}$ is the original denoised guard prediction. Therefore, the guard-frame correction is bounded by a fixed ratio of the local denoised prediction energy. This explains why RCSS can improve boundary continuation without applying unconstrained modifications to the solver state.

\section{Additional Evaluation Details}
\label{app:evaluation_details}

\subsection{Case-Level Aggregation}
\label{app:case_level_aggregation}

The benchmark contains cases with different numbers of prompts. Some cases contain two prompts, while others contain three or more prompts. Directly averaging over all generated segments would give larger weights to cases with more prompts. Therefore, for segment-level metrics, we first average the scores within each case and then average over all cases:
\begin{equation}
S_{\mathrm{case}}
=
\frac{1}{|\mathcal{C}|}
\sum_{c\in\mathcal{C}}
\left(
\frac{1}{N_c}
\sum_{j=1}^{N_c}
S_{c,j}
\right),
\label{eq:case_level_aggregation}
\end{equation}
where $\mathcal{C}$ is the set of benchmark cases, $N_c$ is the number of evaluated segments in case $c$, and $S_{c,j}$ is the score of the $j$-th segment. This aggregation ensures that each case contributes equally regardless of the number of prompts.

\subsection{Prompt-Level Text Alignment}
\label{app:text_alignment}

For Text-Image Similarity~\cite{radford2021learning}, all videos are generated using extended prompts, while
the evaluation references differ across experiments. Tables~\ref{tab:main_quantitative_results}, \ref{tab:ablation_results}, \ref{tab:per_seed_results_all_methods}, and~\ref{tab:paired_significance} use the corresponding short prompts, whereas the remaining experiments use extended prompts. Each segment is evaluated against its corresponding prompt or transition description rather than the full prompt sequence. Consequently, absolute similarity scores across these settings are not directly comparable, while comparisons within each experiment follow a consistent evaluation protocol.

\subsection{Random Seeds}
\label{app:random_seeds}

All quantitative results in the main paper are averaged over three random seeds. This reduces the influence of favorable or unfavorable random initialization and provides a more stable estimate of each method's performance. The same seed set is used for all compared methods whenever the baseline implementation supports seed control.

\subsection{Blind Human Study Protocol}
\label{app:human_protocol}

Our blind study includes 30 participants and 40 anonymized cases covering all ten prompt-transition modes and three random seeds. We evaluate MEVG~\cite{oh2024mevg}, Video-Infinity~\cite{tan2024video}, DiTCtrl~\cite{cai2025ditctrl}, and \textsc{FreqSync} using five dimensions, each scored on a five-point scale: (i) target semantic correctness, (ii) source leakage and artifacts, (iii) transition smoothness, (iv) visual quality, and (v) overall preference. For criterion (ii), a higher score means fewer unwanted source remnants, ghosting, duplicated subjects, or artifacts, so all five reported metrics share the same higher-is-better direction. Table~\ref{tab:human_study} reports the mean ratings. We treat this study as complementary perceptual evidence rather than a replacement for matched automatic metrics.

\section{Additional Quantitative Results}
\label{app:additional_quantitative}
\label{app:per_seed_results}

Table~\ref{tab:per_seed_results_all_methods} reports the per-seed results of all compared methods under three random seeds, namely 42, 123, and 999. We also report the averaged scores across these three seeds, which are used in the main paper. The average scores are computed from the original unrounded per-seed results, while the displayed per-seed values are rounded for readability. Therefore, recomputing averages directly from the displayed rounded values may lead to minor differences in the last decimal place. This per-seed breakdown provides a more detailed view of the stability of each method across different random initializations.

\begin{table*}[h]
\centering
\caption{Per-seed quantitative results of all compared methods.
We report the results of MEVG~\cite{oh2024mevg}, Video-Infinity~\cite{tan2024video}, DiTCtrl~\cite{cai2025ditctrl}, and \textsc{FreqSync} under three random seeds: 42, 123, and 999. The shaded average rows correspond to the results reported in the main paper. CSCV and Motion Smoothness are reported as percentages.}

\label{tab:per_seed_results_all_methods}
\small
\setlength{\tabcolsep}{5.0pt}
\renewcommand{\arraystretch}{1.12}
\resizebox{\linewidth}{!}{
\begin{tabular}{l|c|cccccc}
\toprule
Method 
& Seed 
& \makecell{CSCV\\(\%)$\uparrow$} 
& \makecell{Motion\\Smooth.\\(\%)$\uparrow$} 
& \makecell{Text-Image\\Similarity$\uparrow$} 
& \makecell{Visual\\Quality$\uparrow$} 
& \makecell{T2V\\Align.$\uparrow$} 
& \makecell{Physical\\Consist.$\uparrow$} \\
\midrule

MEVG~\cite{oh2024mevg} 
& 42  & 68.09 & 95.54 & 29.86 & 1.957 & 2.778 & 2.261 \\
MEVG~\cite{oh2024mevg} 
& 123 & 67.23 & 96.04 & 29.29 & 2.008 & 2.868 & 2.306 \\
MEVG~\cite{oh2024mevg} 
& 999 & 66.34 & 95.39 & 29.52 & 1.946 & 2.674 & 2.272 \\
\rowcolor{gray!10}
MEVG~\cite{oh2024mevg} 
& Average & 67.22 & 95.66 & 29.56 & 1.970 & 2.773 & 2.280 \\

\midrule

Video-Infinity~\cite{tan2024video} 
& 42  & 83.72 & 96.53 & 29.92 & 3.255 & 3.511 & 3.200 \\
Video-Infinity~\cite{tan2024video} 
& 123 & 83.63 & 96.38 & 29.75 & 3.210 & 3.347 & 3.162 \\
Video-Infinity~\cite{tan2024video} 
& 999 & 83.52 & 96.42 & 29.49 & 3.275 & 3.409 & 3.268 \\
\rowcolor{gray!10}
Video-Infinity~\cite{tan2024video} 
& Average & 83.62 & 96.45 & 29.72 & \textbf{3.247} & 3.422 & 3.210 \\

\midrule

DiTCtrl~\cite{cai2025ditctrl} 
& 42  & 86.46 & 98.48 & 30.35 & 3.129 & 3.753 & 3.230 \\
DiTCtrl~\cite{cai2025ditctrl} 
& 123 & 87.45 & 98.59 & 30.48 & 3.171 & 3.801 & 3.275 \\
DiTCtrl~\cite{cai2025ditctrl} 
& 999 & 84.42 & 98.25 & 30.14 & 3.142 & 3.721 & 3.181 \\
\rowcolor{gray!10}
DiTCtrl~\cite{cai2025ditctrl} 
& Average & \underline{86.11} & \textbf{98.44} & \underline{30.33} & 3.147 & \underline{3.758} & \underline{3.228} \\

\midrule

\textsc{FreqSync} (Ours) 
& 42  & 87.39 & 98.19 & 30.50 & 3.244 & 3.864 & 3.327 \\
\textsc{FreqSync} (Ours) 
& 123 & 87.00 & 98.42 & 30.58 & 3.203 & 3.854 & 3.191 \\
\textsc{FreqSync} (Ours) 
& 999 & 84.91 & 98.15 & 30.51 & 3.195 & 3.829 & 3.203 \\
\rowcolor{gray!12}
\textsc{FreqSync} (Ours) 
& Average & \textbf{86.43} & \underline{98.25} & \textbf{30.53} & \underline{3.214} & \textbf{3.849} & \textbf{3.240} \\

\bottomrule
\end{tabular}}
\end{table*}

\section{Paired Statistical Significance}
\label{app:statistical_significance}

To assess the reliability of improvements over DiTCtrl~\cite{cai2025ditctrl}, we evaluate 100 balanced cases covering all ten transition modes, matched on prompts, cases, random seed, and generation protocol. We form stratified paired-bootstrap 95\% confidence intervals (CIs), apply case-level sign-flip tests, and correct the resulting $p$-values with the Holm procedure. Table~\ref{tab:paired_significance} reports five metrics for which the corrected comparisons are available. The CSCV gain is +1.78~pp with a 95\% CI of $[+0.86,+2.78]$~pp and Holm-adjusted $p=0.00124$. All five bootstrap confidence intervals exclude zero, and the corresponding Holm-adjusted $p$-values are below 0.05.

\begin{table}[t]
\centering
\small
\setlength{\tabcolsep}{4pt}
\renewcommand{\arraystretch}{1.10}
\caption{Paired 100-case analysis against DiTCtrl. Differences are FreqSync minus DiTCtrl; CSCV differences are in percentage points (pp). All listed confidence intervals exclude zero. Minor differences from subtracting displayed scores arise from rounding.}
\label{tab:paired_significance}
\resizebox{\linewidth}{!}{%
\begin{tabular}{lccccc}
\toprule
Metric & FreqSync & DiTCtrl & Difference & 95\% CI & Holm $p$ \\
\midrule
CSCV (\%) & \textbf{87.5133} & 85.7334 & +1.7799 & [0.8623, 2.7845] & 0.00124 \\
Text--Image Similarity & \textbf{30.5638} & 30.3236 & +0.2402 & [0.0244, 0.4653] & 0.04325 \\
Visual Quality & \textbf{3.2980} & 3.0855 & +0.2124 & [0.1536, 0.2725] & $<10^{-4}$ \\
T2V Alignment & \textbf{3.9049} & 3.7341 & +0.1708 & [0.1003, 0.2429] & 0.00012 \\
Physical Consistency & \textbf{3.3885} & 3.1571 & +0.2314 & [0.1396, 0.3251] & $<10^{-4}$ \\
\bottomrule
\end{tabular}}
\end{table}

\section{Cross-Backbone Evaluation}
\label{app:cross_backbone}

We extend the evaluation to two pre-trained video generators: Wan2.1-1.3B~\cite{wan2025wan}, based on DiT, and VideoCrafter2~\cite{chen2024videocrafter2}, based on a U-Net. Within each backbone, DiTCtrl~\cite{cai2025ditctrl} and \textsc{FreqSync} use the same cases, seed, checkpoint, resolution, sampling steps, classifier-free guidance, window length, and overlap. This comparison tests transfer beyond CogVideoX-2B.

\begin{table*}[t]
\centering
\small
\setlength{\tabcolsep}{5pt}
\renewcommand{\arraystretch}{1.10}
\caption{Cross-backbone evaluation. All metrics are higher-is-better; CSCV and Motion Smoothness are percentages. Best results are bold within each backbone.}
\label{tab:cross_backbone}
\resizebox{\textwidth}{!}{%
\begin{tabular}{llcccccc}
\toprule
Backbone & Method & CSCV (\%) & Motion (\%) & Text--Image & VQ & T2V & Physical \\
\midrule
VideoCrafter2 & DiTCtrl & 81.04 & 97.15 & 34.01 & \textbf{3.095} & \textbf{3.665} & \textbf{3.057} \\
VideoCrafter2 & \textsc{FreqSync} & \textbf{86.50} & \textbf{99.35} & \textbf{34.23} & 2.836 & 3.576 & 2.842 \\
\midrule
Wan2.1-1.3B & DiTCtrl & 84.05 & \textbf{98.25} & 31.91 & 3.267 & 3.900 & 3.404 \\
Wan2.1-1.3B & \textsc{FreqSync} & \textbf{87.00} & 98.04 & \textbf{33.50} & \textbf{3.872} & \textbf{4.095} & \textbf{3.809} \\
\bottomrule
\end{tabular}}
\vspace{-0.6em}
\end{table*}

As shown in Table~\ref{tab:cross_backbone}, CSCV improves by 5.46~pp on VideoCrafter2 (95\% CI $[3.85,7.13]$~pp) and by 2.94~pp on Wan2.1 (95\% CI $[0.27,5.50]$~pp). On Wan2.1, \textsc{FreqSync} improves five of six metrics. On VideoCrafter2, boundary continuity, motion smoothness, and text--image alignment improve, while the three perceptual-quality metrics decrease. The factorized U-Net pathways and shorter four-frame overlap in this setting require stronger temporal-memory and stitching corrections, which may suppress details while improving continuity. The cross-backbone findings therefore support transfer of the boundary-coupling principle while also exposing architecture-dependent trade-offs.

\section{Controlled Frequency-Routing Ablations}
\label{app:routing_ablation}

To test the role of frequency allocation directly, we run six controlled variants on the same 40 cases, spanning all ten transition modes with matched prompts, seed, generation protocol, and evaluation. In Table~\ref{tab:controlled_routing}, $(L,M,H)$ denotes the source-conditioned routing strengths applied to low-, mid-, and high-frequency components of FS-SCA. Setting a source weight to zero removes that \emph{source correction}, not the corresponding native target-frequency component. Accordingly, ``high-only'' is not a removal of low-frequency information from the target branch.

\begin{table*}[t]
\centering
\small
\setlength{\tabcolsep}{5pt}
\renewcommand{\arraystretch}{1.10}
\caption{Controlled 40-case frequency-routing ablations. Mean $\pm$ standard deviation; all scores are higher-is-better. The routing tuple gives source weights $(L,M,H)$.}
\label{tab:controlled_routing}
\resizebox{\textwidth}{!}{%
\begin{tabular}{lcccc}
\toprule
Variant & Source routing $(L,M,H)$ & CSCV (\%) & Text--Image Similarity & T2V Alignment \\
\midrule
Full \textsc{FreqSync} & $(0.98,0.68,0.035)$ & $\mathbf{89.139\pm6.774}$ & $34.013\pm3.019$ & $\mathbf{3.951\pm0.370}$ \\
Low+Mid-only & $(0.98,0.68,0)$ & $88.956\pm7.121$ & $\mathbf{34.027\pm3.028}$ & $3.891\pm0.369$ \\
High-only & $(0,0,0.98)$ & $86.302\pm7.335$ & $33.913\pm3.064$ & $3.805\pm0.418$ \\
Reverse Routing & $(0.035,0.68,0.98)$ & $85.795\pm8.250$ & $33.885\pm2.939$ & $3.821\pm0.349$ \\
Uniform Routing & $(0.565,0.565,0.565)$ & $88.380\pm7.506$ & $33.964\pm3.075$ & $3.903\pm0.411$ \\
w/o FS-SCA & --- & $85.693\pm8.385$ & $33.570\pm2.902$ & $3.905\pm0.361$ \\
\bottomrule
\end{tabular}}
\vspace{-1.2em}
\end{table*}

Low+Mid-only nearly matches the full method, while High-only and Reverse Routing lower CSCV by 2.84 and 3.34~pp, respectively. Uniform Routing controls for mean source strength but lowers all three metrics relative to the full model. Removing FS-SCA decreases CSCV by 3.446~pp and Text--Image Similarity by 0.443. This controlled intervention supports asymmetric source-low/mid and target-high routing rather than an explanation based only on stronger uniform coupling. It also extends the qualitative boundary diagnosis beyond a single failure example.

\section{Hyperparameter Sensitivity}
\label{app:sensitivity_analysis}

We assess local robustness around the fixed default configuration with a paired one-factor-at-a-time (OFAT) sweep and a separate cutoff--layer interaction study. The same default configuration is used across cases within each study, without per-case tuning. For the short temporal Fourier axis, the sweep logs describe selected partitions using the discrete ``DC + non-DC bins'' convention; the implementation settings in Sec.~\ref{subsec:experimental_setup} report normalized cutoffs. Each study is evaluated under its corresponding experimental protocol, and comparisons are made within the same setting.

\begin{table*}[t]
\centering
\scriptsize
\setlength{\tabcolsep}{3pt}
\renewcommand{\arraystretch}{1.10}
\caption{Paired OFAT sensitivity analysis. Absolute scores are shown for each configuration; the default is the reference. ``DC+$k$'' is the logged temporal-frequency-bin partition. No tested alternative dominates the default across all metrics.}
\label{tab:sensitivity_ofat}
\resizebox{\textwidth}{!}{%
\begin{tabular}{llcccccc}
\toprule
Family & Setting & CSCV (\%) & Text--Image & VQ & T2V & Physical & Motion (\%) \\
\midrule
Default & low=2, low+mid=3, 8 layers & 88.993 & 34.083 & 3.395 & 3.984 & 3.419 & 98.184 \\
Low cutoff & DC+$1$ non-DC & 89.326 & 34.161 & 3.414 & 3.987 & 3.377 & 98.191 \\
Low cutoff & DC+$3$ non-DC & 88.890 & 34.092 & 3.410 & 3.960 & 3.453 & 98.183 \\
Low+mid cutoff & DC+$2$, mid collapsed & 88.625 & 34.112 & 3.300 & 3.862 & 3.295 & 98.190 \\
Low+mid cutoff & DC+$4$ & 88.845 & 34.078 & 3.311 & 3.925 & 3.352 & 98.182 \\
Routing & $0.8\times$ & 88.315 & 34.123 & 3.298 & 4.014 & 3.302 & 98.191 \\
Routing & $1.2\times$ & 89.228 & 34.100 & 3.324 & 3.926 & 3.323 & 98.184 \\
Active layers & $[20,30)$, 10 layers & 88.069 & 34.108 & 3.425 & 4.023 & 3.338 & 98.161 \\
Active layers & $[24,30)$, 6 layers & 88.967 & 34.082 & 3.333 & 3.865 & 3.445 & 98.206 \\
\bottomrule
\end{tabular}}
\end{table*}

\begin{table}[t]
\centering
\small
\setlength{\tabcolsep}{5pt}
\renewcommand{\arraystretch}{1.10}
\caption{Cutoff--layer interaction sweep. Every cell lists CSCV (\%)/T2V Alignment. These interaction results come from a separate paired study and should not be combined numerically with the OFAT table.}
\label{tab:sensitivity_interaction}
\begin{tabular}{lccc}
\toprule
Low partition & 10 layers $[20,30)$ & 8 layers $[22,30)$ & 6 layers $[24,30)$ \\
\midrule
DC+$1$ & 87.514 / 4.000 & 88.476 / 3.799 & 88.739 / 3.846 \\
DC+$2$ & 87.000 / 3.948 & 88.289 / 3.918 & 89.494 / 3.866 \\
DC+$3$ & 87.491 / 4.073 & 88.070 / 3.694 & 88.697 / 3.859 \\
\bottomrule
\end{tabular}
\vspace{-1.5em}
\end{table}

Tables~\ref{tab:sensitivity_ofat} and~\ref{tab:sensitivity_interaction} show that \textsc{FreqSync} maintains stable performance under variations in frequency cutoffs, routing strengths, and active layers, with no alternative configuration consistently outperforming the default. All cutoff--layer interaction 95\% CIs include zero, indicating no statistically significant interaction or evidence of fragile parameter dependence. The three-band design provides a balanced performance across metrics, while only the $0.8\times$ routing setting shows a mild degradation on fast-motion cases. Overall, \textsc{FreqSync} is not highly sensitive to these hyperparameters around the default configuration and requires no per-case tuning.

\section{Inference Efficiency}
\label{app:efficiency_analysis}

We compare \textsc{FreqSync} with DiTCtrl~\cite{cai2025ditctrl} on the same NVIDIA A800-SXM4-80GB GPU using CogVideoX-2B, $480\times720$ resolution, 50 denoising steps, and matched generation settings. Each 2--5-prompt sequence receives one warm-up and three CUDA-synchronized timed runs. We time the core latent-sampling interval and exclude shared model loading, text encoding, VAE decoding, and video I/O. FLOPs are counted using one multiply--accumulate as two floating-point operations; reported memory is peak allocated GPU memory. This protocol compares the proposed sampling components under controlled conditions, not end-to-end wall-clock video-delivery latency.

\begin{table*}[t]
\centering
\scriptsize
\setlength{\tabcolsep}{3pt}
\renewcommand{\arraystretch}{1.12}
\caption{Inference cost as prompt count increases. Runtime is mean $\pm$ standard deviation over three timed runs (seconds); memory is peak allocated GiB. Counted FLOPs are in TFLOPs per sequence. Percentage increases are relative to DiTCtrl.}
\label{tab:inference_cost}
\resizebox{\textwidth}{!}{%
\begin{tabular}{c|ccc|ccc|ccc}
\toprule
\multirow{2}{*}{Prompts} & \multicolumn{3}{c|}{Sampling time (s)} & \multicolumn{3}{c|}{Memory (GiB)} & \multicolumn{3}{c}{Counted TFLOPs} \\
& DiTCtrl & FreqSync & $\Delta$ & DiTCtrl & FreqSync & $\Delta$ & DiTCtrl & FreqSync & $\Delta$ \\
\midrule
2 & $461.05\pm0.28$ & $502.85\pm2.41$ & +9.07\% & 16.51 & 16.74 & +1.39\% & 71,998.39 & 75,379.93 & +4.697\% \\
3 & $923.07\pm0.39$ & $1000.50\pm0.69$ & +8.39\% & 16.53 & 16.76 & +1.39\% & 143,996.77 & 150,759.87 & +4.697\% \\
4 & $1383.91\pm0.47$ & $1500.52\pm1.76$ & +8.43\% & 16.56 & 16.79 & +1.37\% & 215,995.16 & 226,139.80 & +4.697\% \\
5 & $1851.45\pm0.92$ & $2000.84\pm0.97$ & +8.07\% & 16.59 & 16.81 & +1.37\% & 287,993.55 & 301,519.73 & +4.697\% \\
\bottomrule
\end{tabular}}
\end{table*}

As shown in Table~\ref{tab:inference_cost}, peak allocated GPU memory increases by only 0.23 GiB (233 MiB, 1.37\%). Across 2--5 prompts, the sampling-time overhead remains within 8.07--9.07\%, while the counted-FLOP overhead stays at 4.697\% and peak memory remains nearly constant. Overall, \textsc{FreqSync} achieves improved generation quality with modest and stable inference overhead, without introducing additional trainable parameters or training costs.

\section{Additional Diagnostic Visualizations}
\label{app:additional_diagnostic}

\begin{figure*}[t]
    \centering
    \includegraphics[width=0.88\linewidth]{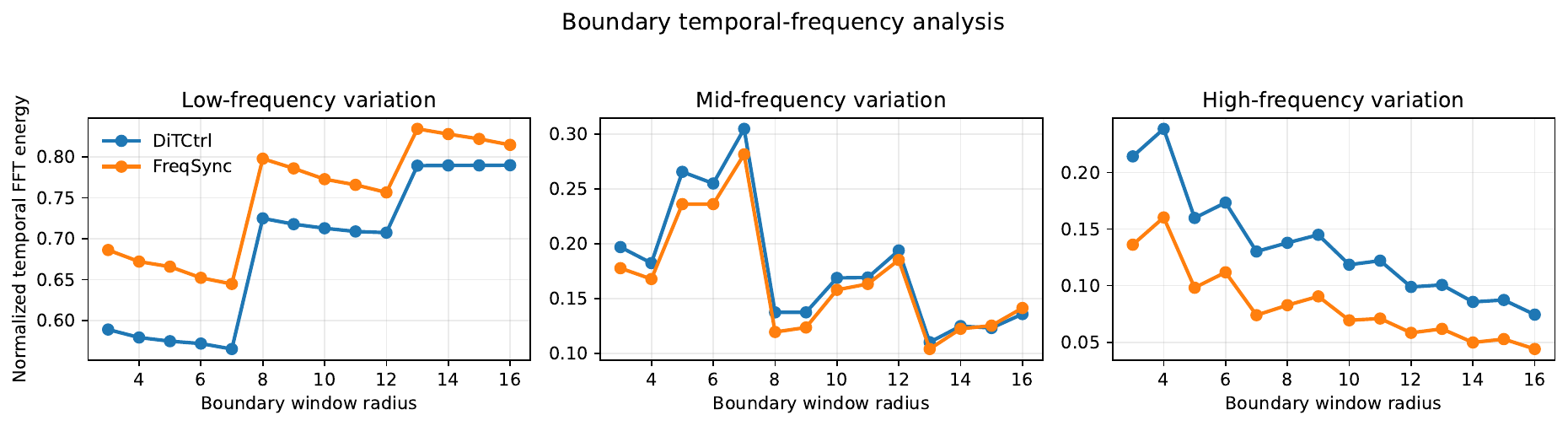}
    \caption{Boundary temporal-frequency analysis. \textsc{FreqSync} shows stronger low-frequency continuation and lower high-frequency variation around prompt boundaries. This supports our frequency-synchronized boundary coupling design, where low-frequency components preserve structural continuity and high-frequency instability is suppressed near transitions.}
    \label{fig:supp_temporal_frequency_energy}
\end{figure*}

We also analyze temporal-frequency energy around prompt boundaries. Figure~\ref{fig:supp_temporal_frequency_energy} reports normalized temporal FFT energy under different boundary window radii. \textsc{FreqSync} exhibits stronger low-frequency energy near the boundary, which indicates more stable structural and motion continuation. Meanwhile, its high-frequency variation is consistently lower than DiTCtrl~\cite{cai2025ditctrl}, suggesting that the transition contains fewer abrupt high-frequency changes. This behavior is consistent with the design of FS-SCA and RCSS: low- and mid-frequency components are used to carry continuity cues, while high-frequency components remain target-dominant and are constrained to reduce boundary-local flicker, ghosting, and duplicated traces.

\section{Additional Qualitative Visualizations}
\label{app:additional_qualitative}

The main paper presents a representative qualitative comparison and a visual ablation study. In this section, we provide additional qualitative visualizations from three perspectives. First, we show more generated examples of \textsc{FreqSync} under diverse multi-prompt sequences and transition types. Second, we present a single-prompt long-video generation application, where the same prompt is assigned to all segments and \textsc{FreqSync} is used to extend the video duration. Third, we provide additional qualitative comparisons with Kling-2.6~\cite{kling2024official}, MEVG~\cite{oh2024mevg}, Video-Infinity~\cite{tan2024video}, and DiTCtrl~\cite{cai2025ditctrl} on challenging three-prompt sequences.

\subsection{Additional Generated Examples}
\label{app:additional_generated_examples}

\begin{figure}[!htbp]
    \centering
    \includegraphics[width=0.96\linewidth]{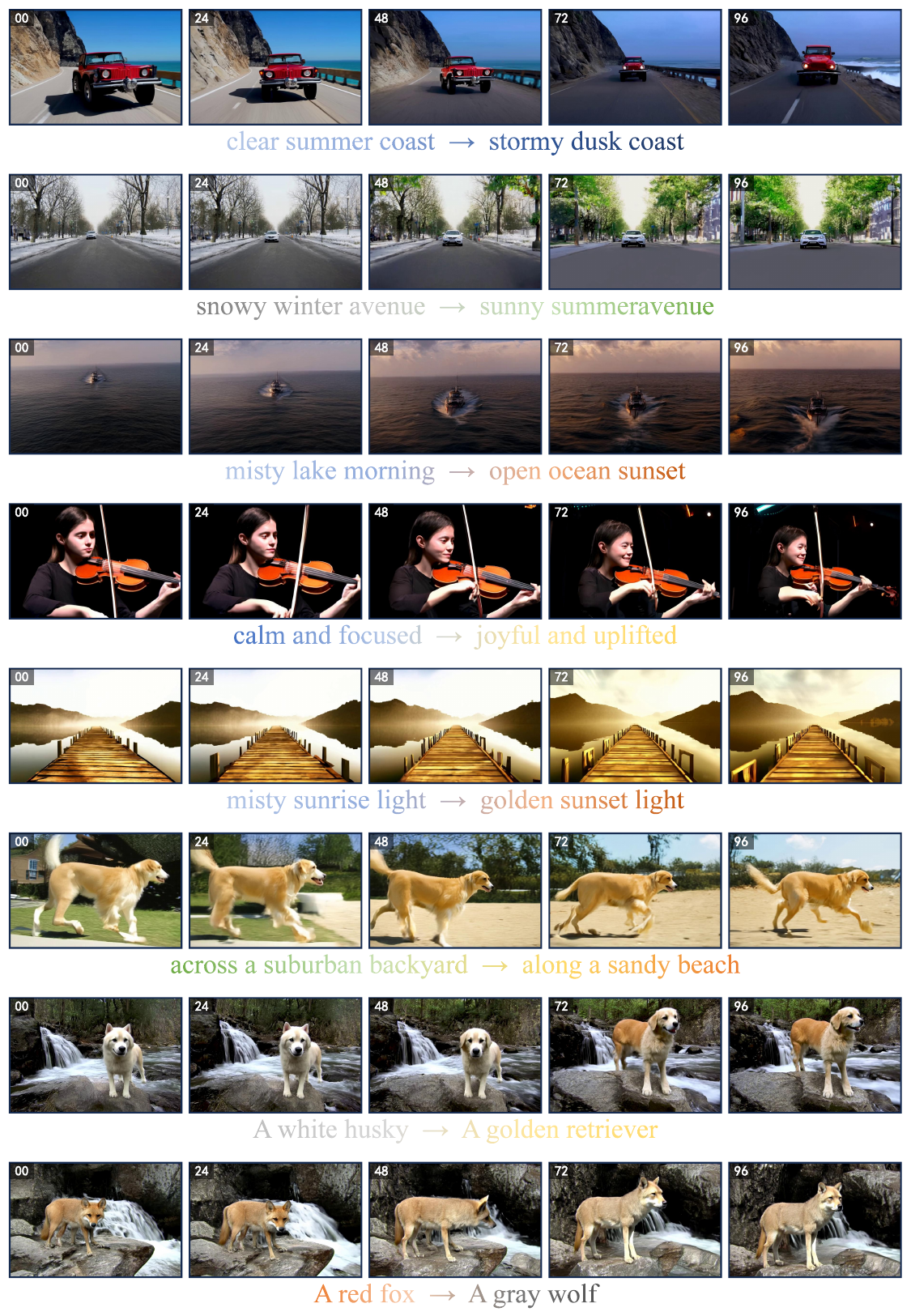}
    \vspace{-0.6em}
    \caption{Additional two-prompt generated examples.
    We show eight representative two-prompt sequences. Each row corresponds to one generated video and visualizes key frames across a single prompt boundary. Across diverse transition types, \textsc{FreqSync} preserves the main subject, scene layout, and motion trajectory while allowing the target-prompt condition to emerge progressively. These examples further illustrate that the frequency-synchronized boundary coupling used by FS-SCA and RCSS can reduce abrupt visual changes without suppressing prompt-specific details.}
    \label{fig:supp_two_prompt_examples}
    \vspace{-0.6em}
\end{figure}

\begin{figure}[!htbp]
    \centering
    \includegraphics[width=0.96\linewidth]{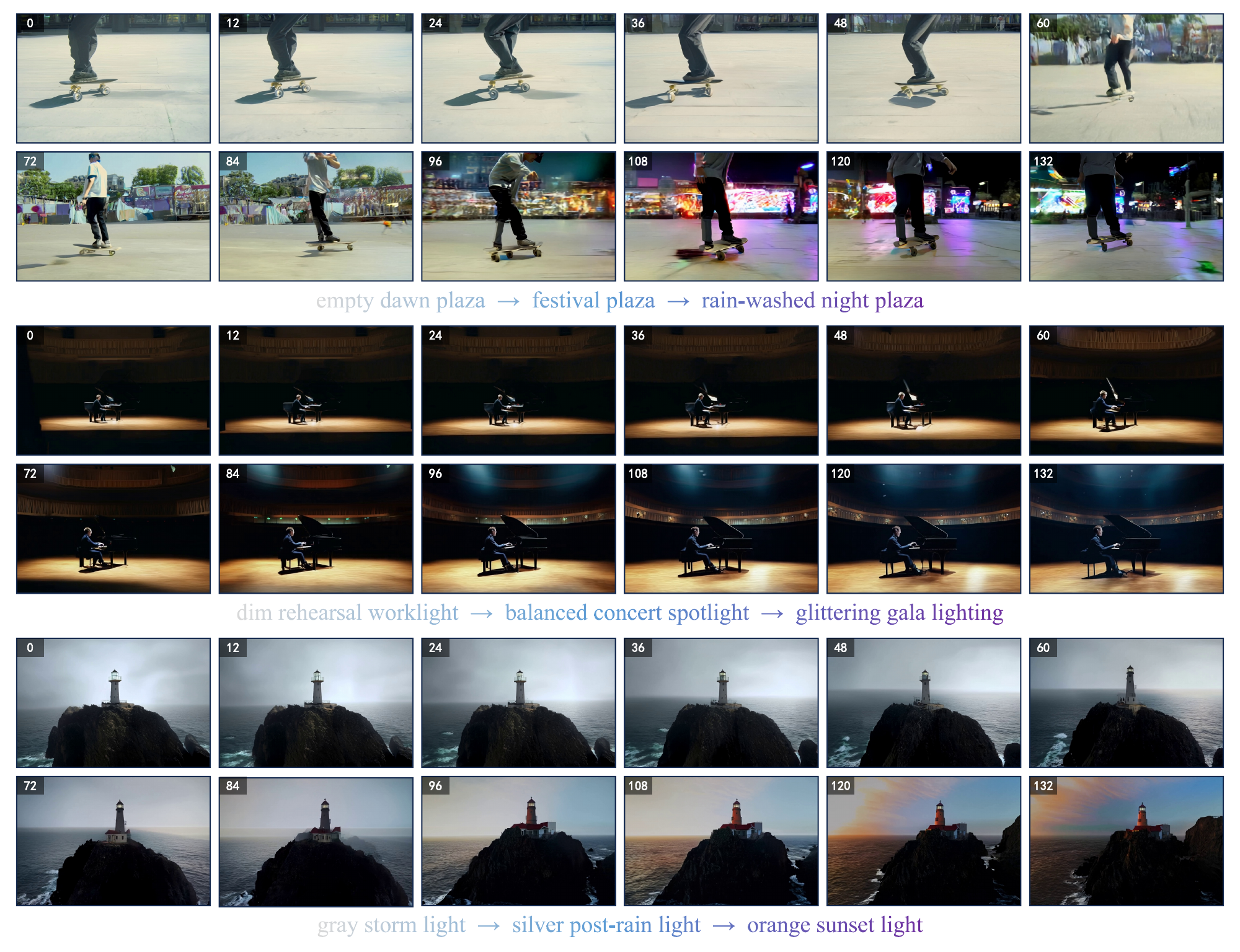}
    \caption{Additional three-prompt generated examples.
    We show three representative three-prompt sequences, where each example spans two rows to cover a longer temporal range and two consecutive prompt boundaries. \textsc{FreqSync} maintains coherent temporal evolution across multiple semantic changes, such as scene, illumination, and action transitions. The results suggest that the proposed boundary coupling is not limited to a single transition, but can also support longer prompt sequences with successive target-prompt activations.}
    \label{fig:supp_three_prompt_examples}
\end{figure}

Figure~\ref{fig:supp_two_prompt_examples} shows two-prompt examples with a single prompt boundary, while Figure~\ref{fig:supp_three_prompt_examples} shows three-prompt examples with two consecutive prompt boundaries. These examples cover diverse changes in background, lighting, location, subject, and action conditions. For each sequence, we visualize uniformly sampled frames together with the corresponding prompt transition labels. The results further show that \textsc{FreqSync} can preserve temporal continuity while allowing target-prompt details to emerge progressively and remain visually coherent throughout the sequence.

\FloatBarrier

\subsection{Single-Prompt Long-Video Generation}
\label{app:single_prompt_generation}

\begin{figure*}[p]
    \centering
    \includegraphics[width=0.88\linewidth]{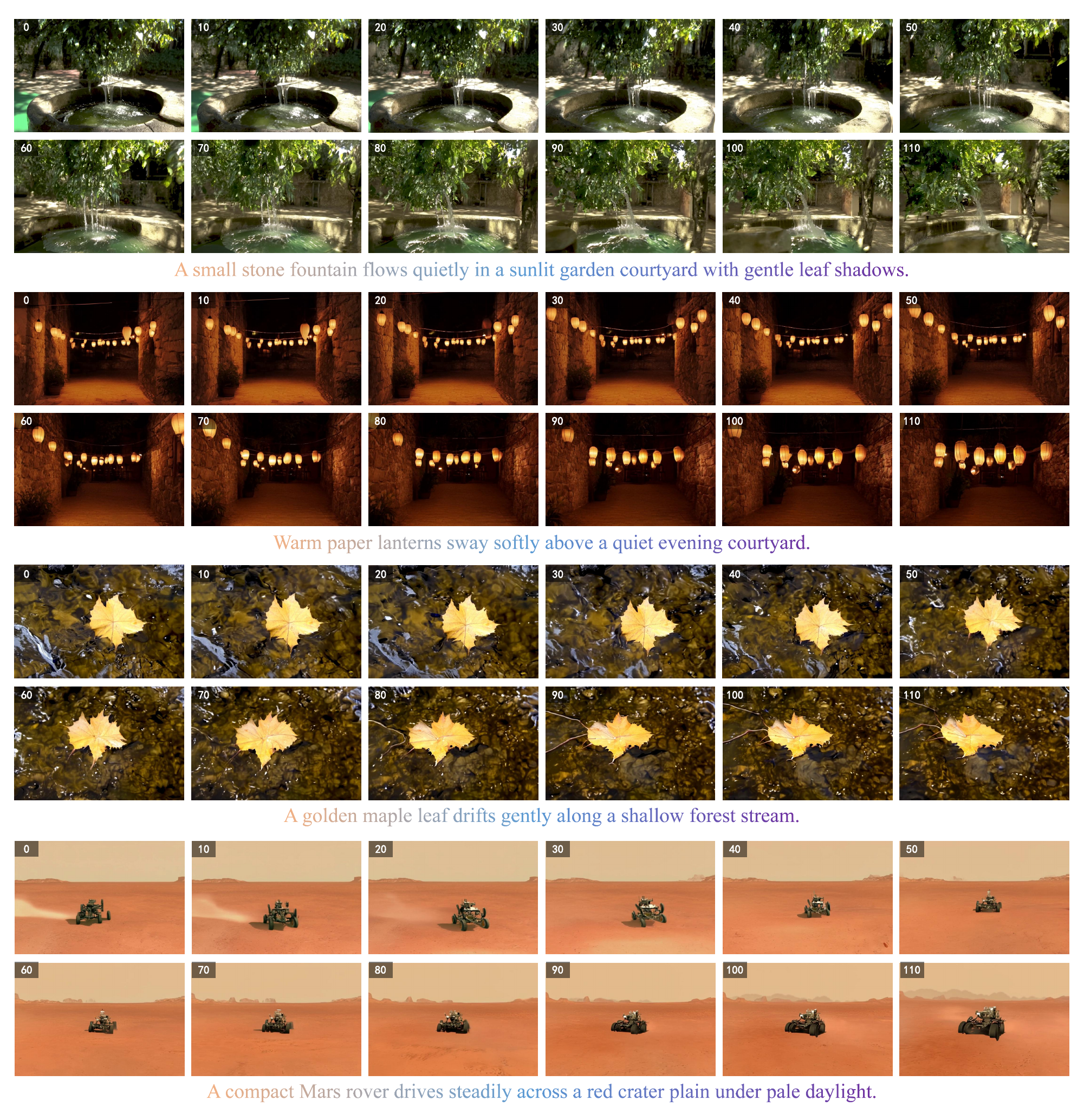}
    \caption{Single-prompt long-video generation with \textsc{FreqSync}.
    We apply \textsc{FreqSync} to single-prompt long-video generation by assigning the same prompt to all segments. The four examples show a sunlit courtyard fountain, warm paper lanterns, a drifting maple leaf, and a Mars rover driving across a red crater plain. Even without semantic prompt changes, long-video generation still requires stable cross-segment continuation. \textsc{FreqSync} produces smooth motion and coherent visual content over extended durations, suggesting that its boundary-coupling mechanism can also benefit single-prompt video extension.}
    \label{fig:supp_single_prompt_generation}
\end{figure*}

Although \textsc{FreqSync} is designed for multi-prompt long video generation, it can also be applied to single-prompt long-video generation by assigning the same prompt to every segment. In this setting, there is no semantic prompt transition, but adjacent segments still need to be connected smoothly to avoid temporal seams, flicker, or layout drift. Figure~\ref{fig:supp_single_prompt_generation} shows four single-prompt long-video examples generated by \textsc{FreqSync}. The results demonstrate that the proposed boundary coupling can also support temporally coherent long-video extension under a fixed prompt, maintaining stable scene content and smooth motion across segment boundaries.

\subsection{Additional Qualitative Comparisons}
\label{app:additional_qualitative_comparisons}

\begin{figure*}[p]
    \centering
    \includegraphics[width=\linewidth]{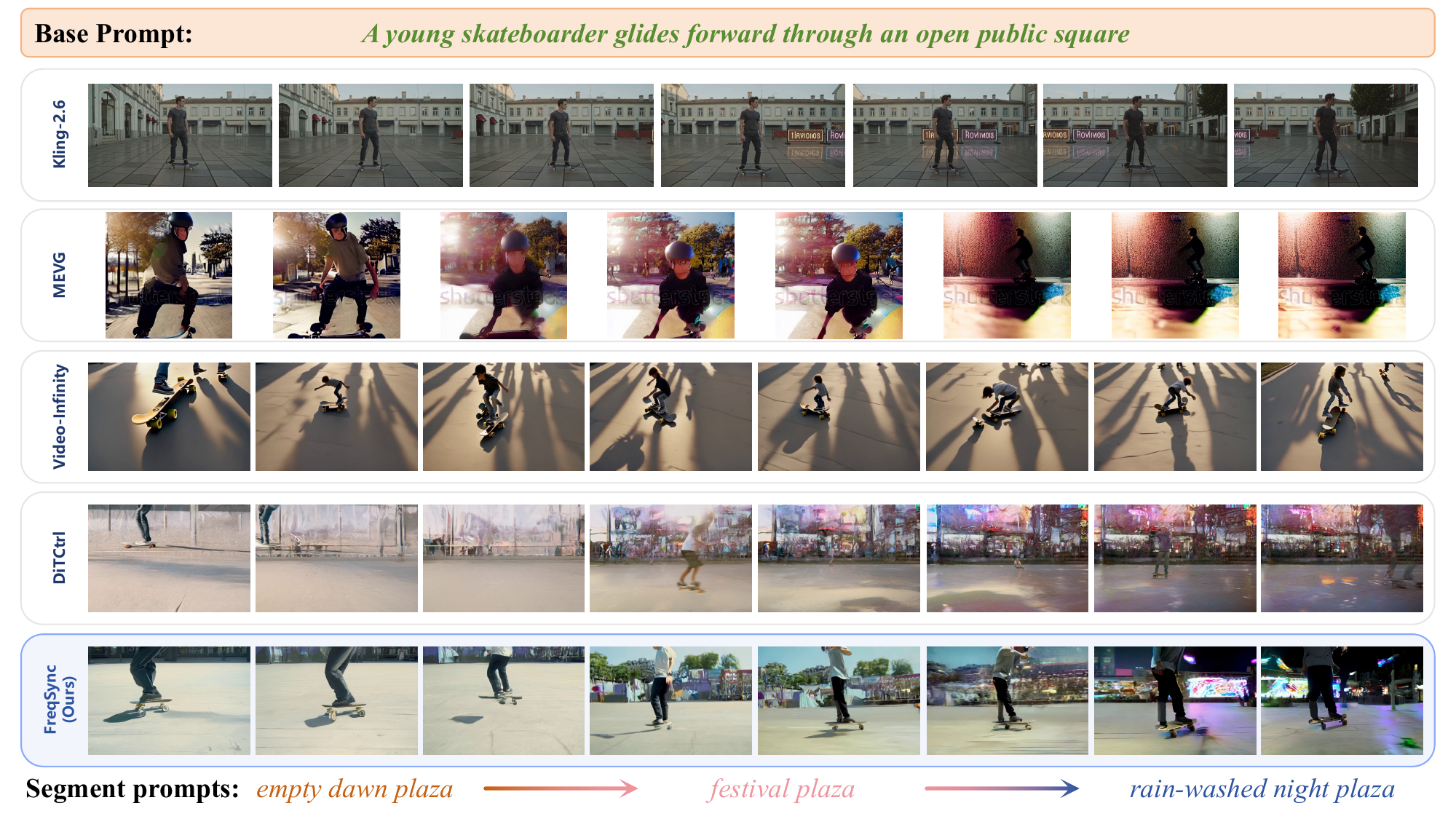}
    \caption{Additional qualitative comparison on a skateboarder sequence.
    The base prompt describes a young skateboarder gliding forward through an open public square, while the segment prompts transition from an empty dawn plaza to a festival plaza and then to a rain-washed night plaza. Kling-2.6~\cite{kling2024official} produces visually stable frames but shows limited scene progression. MEVG~\cite{oh2024mevg} and Video-Infinity~\cite{tan2024video} suffer from identity, viewpoint, or motion inconsistency. DiTCtrl~\cite{cai2025ditctrl} improves cross-segment continuity but introduces ghosting and blurred boundary artifacts. In contrast, \textsc{FreqSync} preserves the skateboarder's motion trajectory while progressively changing the plaza environment and illumination.}
    \label{fig:supp_comparison_skateboard}
\end{figure*}

\begin{figure*}[p]
    \centering
    \includegraphics[width=\linewidth]{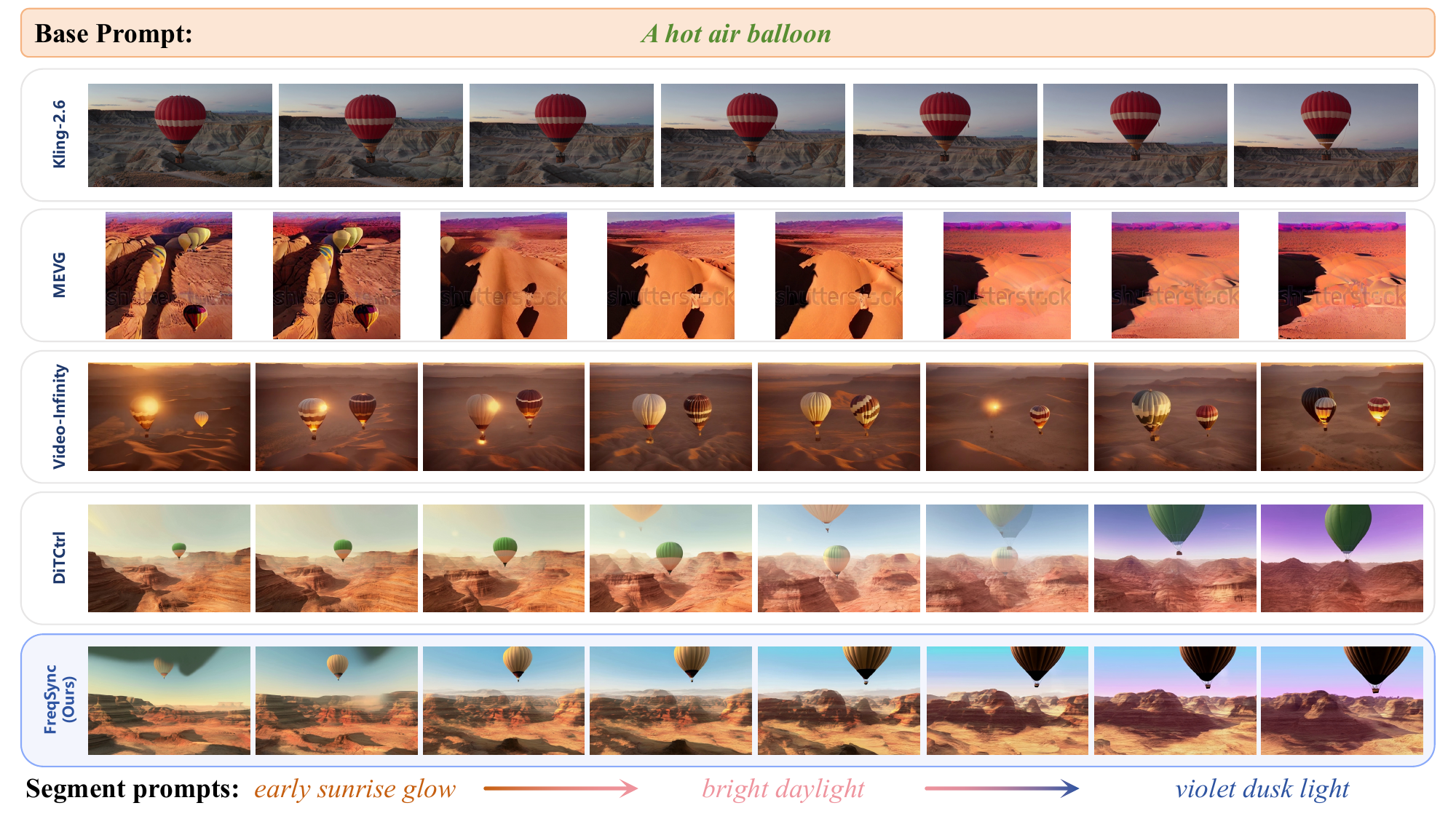}
    \caption{Additional qualitative comparison on a hot-air-balloon sequence.
    The base prompt describes a hot air balloon, and the segment prompts change the lighting condition from early sunrise glow to bright daylight and then to violet dusk light. Kling-2.6~\cite{kling2024official} preserves the balloon appearance but shows limited temporal and illumination transition. MEVG~\cite{oh2024mevg} and Video-Infinity~\cite{tan2024video} produce noticeable subject or scene inconsistencies, including multiple balloons, unstable scale, or abrupt viewpoint changes. DiTCtrl~\cite{cai2025ditctrl} maintains partial continuity but shows over-preserved source-side evidence and boundary-local artifacts. \textsc{FreqSync} better preserves the balloon identity and desert layout while producing a smoother lighting transition across the prompt sequence.}
    \label{fig:supp_comparison_balloon}
\end{figure*}

We further provide two additional qualitative comparisons against Kling-2.6~\cite{kling2024official}, MEVG~\cite{oh2024mevg}, Video-Infinity~\cite{tan2024video}, and DiTCtrl~\cite{cai2025ditctrl}. As shown in Figures~\ref{fig:supp_comparison_skateboard} and~\ref{fig:supp_comparison_balloon}, the compared methods often exhibit weak prompt progression, inconsistent subject appearance, duplicated visual traces, or abrupt boundary changes. In contrast, \textsc{FreqSync} maintains more stable temporal trajectories and smoother prompt transitions while preserving target-prompt semantics.

\section{Broader Impacts}
\label{app:broader_impacts}

\textsc{FreqSync} aims to improve training-free multi-prompt long video generation by producing smoother prompt transitions and more coherent long videos. This may benefit creative video production, storyboarding, visual prototyping, education, and research on controllable video generation, especially because the method does not require additional model training. At the same time, improvements in text-to-video generation may also introduce potential risks. More coherent long videos could be misused to create misleading synthetic media, fabricated events, or deceptive visual narratives. Since \textsc{FreqSync} builds on pre-trained video generators, it may also inherit biases, unsafe behaviors, or inappropriate generation tendencies from the underlying model and its training data. In addition, when generated videos depict people, locations, or sensitive events, misuse may raise concerns related to misinformation, impersonation, privacy, and public trust. We emphasize that \textsc{FreqSync} is a research method for studying prompt-boundary coupling rather than a deployed generation system. We recommend that practical deployments combine such methods with standard safety filters, watermarking or provenance mechanisms, content moderation, and usage policies that restrict malicious applications. We also encourage users to clearly disclose synthetic content when generated videos are used in public-facing contexts.

\section{Licenses and Existing Assets}
\label{app:licenses}

We use several existing models, codebases, benchmarks, and evaluation tools in this work. All external assets are used only for research and evaluation purposes, and we cite their original papers or official releases in the main text. We do not redistribute third-party model weights, benchmark code, or evaluation packages as part of this submission. When using existing assets, we follow their official licenses and terms of use.

\begin{table}[h]
\centering
\caption{Existing assets used in this work. We cite the original sources in the main paper and follow their corresponding licenses or usage terms.}
\label{tab:asset_licenses}
\resizebox{\linewidth}{!}{
\begin{tabular}{lll}
\toprule
Asset & Usage in this work & License / Terms \\
\midrule
CogVideoX-2B~\cite{yang2024cogvideox} 
& Pre-trained video generation backbone 
& Apache-2.0 License \\

DiTCtrl / MPVBench~\cite{cai2025ditctrl} 
& Baseline method and multi-prompt evaluation setting 
& Apache-2.0 License for released inference code \\

VBench~\cite{huang2024vbench} 
& Motion Smoothness evaluation 
& Apache-2.0 License \\

CLIP~\cite{radford2021learning} 
& Text-Image Similarity and CSCV-related feature similarity 
& MIT License \\

VideoScore2~\cite{he2025videoscore2} 
& Visual Quality, T2V Alignment, and Physical Consistency evaluation 
& MIT License \\

MEVG~\cite{oh2024mevg} 
& Baseline comparison 
& Original paper and official project terms \\

Video-Infinity~\cite{tan2024video} 
& Baseline comparison 
& Original paper and official project terms \\

Kling-2.6~\cite{kling2024official} 
& Qualitative comparison with a commercial system 
& Official service terms \\
\bottomrule
\end{tabular}}
\end{table}

\section{Limitations}
\label{app:limitations}

\textsc{FreqSync} has several limitations. First, it relies on the generation capacity of the underlying pre-trained video diffusion model. If the base model fails to synthesize a rare object, complex interaction, or unusual camera motion, \textsc{FreqSync} cannot fully recover the missing content. Second, extremely large semantic changes may still produce imperfect transitions, especially when adjacent prompts require incompatible layouts, object counts, or camera trajectories. Third, FS-SCA introduces additional inference overhead because it evaluates an extra source-conditioned attention path in scheduled layers and denoising steps. Finally, as with other text-to-video generation methods, generated videos may inherit biases or unsafe behaviors from the underlying model and training data. We recommend standard safety filtering and responsible deployment practices for open-ended use.



\clearpage
\section*{NeurIPS Paper Checklist}

\begin{enumerate}

\item {\bf Claims}
    \item[] Question: Do the main claims made in the abstract and introduction accurately reflect the paper's contributions and scope?
    \item[] Answer: \answerYes{} 
    \item[] Justification: The abstract and introduction clearly state the main claims, which accurately reflect the contributions and scope of our work.

\item {\bf Limitations}
    \item[] Question: Does the paper discuss the limitations of the work performed by the authors?
    \item[] Answer: \answerYes{} 
    \item[] Justification: We discuss limitations in Appendix~\ref{app:limitations}.

\item {\bf Theory assumptions and proofs}
    \item[] Question: For each theoretical result, does the paper provide the full set of assumptions and a complete (and correct) proof?
    \item[] Answer: \answerYes{} 
    \item[] Justification: Appendix~\ref{app:bounded_residual} provides the assumptions and proof for the bounded residual correction property used by FS-SCA and RCSS.

    \item {\bf Experimental result reproducibility}
    \item[] Question: Does the paper fully disclose all the information needed to reproduce the main experimental results of the paper to the extent that it affects the main claims and/or conclusions of the paper (regardless of whether the code and data are provided or not)?
    \item[] Answer: \answerYes{} 
    \item[] Justification: We describe the algorithm, inference procedure, benchmark construction, evaluation metrics, and main hyperparameters in Secs.~\ref{sec:method}, \ref{sec:experiments}, and Appendix~\ref{app:algorithmic_details}.

\item {\bf Open access to data and code}
    \item[] Question: Does the paper provide open access to the data and code, with sufficient instructions to faithfully reproduce the main experimental results, as described in supplemental material?
    \item[] Answer: \answerYes{} 
    \item[] Justification: The code and configurations are released at: \url{https://yipanxu-22.github.io/FreqSync-project/}.

\item {\bf Experimental setting/details}
    \item[] Question: Does the paper specify all the training and test details (e.g., data splits, hyperparameters, how they were chosen, type of optimizer) necessary to understand the results?
    \item[] Answer: \answerYes{} 
    \item[] Justification: In Sec.~\ref{subsec:experimental_setup} and Appendix~\ref{app:algorithmic_details}.

\item {\bf Experiment statistical significance}
    \item[] Question: Does the paper report error bars suitably and correctly defined or other appropriate information about the statistical significance of the experiments?
    \item[] Answer: \answerYes{} 
    \item[] Justification: Statistical significance tests and confidence intervals are reported in Appendix~\ref{app:statistical_significance}.

\item {\bf Experiments compute resources}
    \item[] Question: For each experiment, does the paper provide sufficient information on the computer resources (type of compute workers, memory, time of execution) needed to reproduce the experiments?
    \item[] Answer: \answerYes{} 
    \item[] Justification: GPU specifications and inference costs are reported in Sec.~\ref{subsec:experimental_setup} and Appendix~\ref{app:efficiency_analysis}.
    
\item {\bf Code of ethics}
    \item[] Question: Does the research conducted in the paper conform, in every respect, with the NeurIPS Code of Ethics \url{https://neurips.cc/public/EthicsGuidelines}?
    \item[] Answer: \answerYes{} 
    \item[] Justification: We have read and complied with the NeurIPS Code of Ethics.

\item {\bf Broader impacts}
    \item[] Question: Does the paper discuss both potential positive societal impacts and negative societal impacts of the work performed?
    \item[] Answer: \answerYes{} 
    \item[] Justification: We discuss potential positive uses, misuse risks, inherited model biases, and mitigation strategies in Appendix~\ref{app:broader_impacts}.
    
\item {\bf Safeguards}
    \item[] Question: Does the paper describe safeguards that have been put in place for responsible release of data or models that have a high risk for misuse (e.g., pre-trained language models, image generators, or scraped datasets)?
    \item[] Answer: \answerNA{} 
    \item[] Justification: This paper poses no such risks.

\item {\bf Licenses for existing assets}
    \item[] Question: Are the creators or original owners of assets (e.g., code, data, models), used in the paper, properly credited and are the license and terms of use explicitly mentioned and properly respected?
    \item[] Answer: \answerYes{} 
    \item[] Justification: We cite the creators of all models, baselines, benchmarks, and evaluation tools used in our paper, and summarize their licenses or usage terms in Appendix~\ref{app:licenses}.

\item {\bf New assets}
    \item[] Question: Are new assets introduced in the paper well documented and is the documentation provided alongside the assets?
    \item[] Answer: \answerYes{} 
    \item[] Justification: The code and configurations are released at: \url{https://yipanxu-22.github.io/FreqSync-project/}.

\item {\bf Crowdsourcing and research with human subjects}
    \item[] Question: For crowdsourcing experiments and research with human subjects, does the paper include the full text of instructions given to participants and screenshots, if applicable, as well as details about compensation (if any)? 
    \item[] Answer: \answerNo{} 
    \item[] Justification: Full participant instructions and compensation details are not provided, but the human evaluation protocol is described in Sec.~\ref{subsec:human_evaluation} and Appendix~\ref{app:human_protocol}.

\item {\bf Institutional review board (IRB) approvals or equivalent for research with human subjects}
    \item[] Question: Does the paper describe potential risks incurred by study participants, whether such risks were disclosed to the subjects, and whether Institutional Review Board (IRB) approvals (or an equivalent approval/review based on the requirements of your country or institution) were obtained?
    \item[] Answer: \answerNo{} 
    \item[] Justification: Formal IRB approval or equivalent ethics review was not obtained.

\item {\bf Declaration of LLM usage}
    \item[] Question: Does the paper describe the usage of LLMs if it is an important, original, or non-standard component of the core methods in this research? Note that if the LLM is used only for writing, editing, or formatting purposes and does \emph{not} impact the core methodology, scientific rigor, or originality of the research, declaration is not required.
    \item[] Answer: \answerNA{} 
    \item[] Justification: LLMs are not used as an important, original, or non-standard component of the core method or experiments.

\end{enumerate}

\end{document}